\documentclass{article}

\usepackage{microtype}
\usepackage{graphicx}
\usepackage{subfigure}
\usepackage{xspace}
\usepackage[dvipsnames]{xcolor, colortbl}
\usepackage{multirow}
\usepackage{overpic}
\usepackage{caption}
\usepackage{siunitx}
\usepackage{floatrow}
\usepackage{booktabs} %

\usepackage{hyperref}

\usepackage[accepted]{icml2023}

\usepackage{amsmath}
\usepackage{amssymb}
\usepackage{mathtools}
\usepackage{amsthm}

\usepackage[capitalize,noabbrev]{cleveref}

\usepackage{glossaries}
\usepackage[inline]{enumitem}
\glsdisablehyper

\definecolor{TartOrange}{HTML}{ff2e35}
\definecolor{Orange}{HTML}{ff7825}
\definecolor{Mango}{HTML}{ffc013}
\definecolor{AppleGreen}{HTML}{7cb81b}
\definecolor{Blue}{HTML}{1173b0}
\definecolor{BdazzledBlue}{HTML}{2e58a5}
\definecolor{Purple}{HTML}{5b3590}
\definecolor{Sunglow}{HTML}{FFCA3A}
\definecolor{TableRow}{gray}{0.9}

\newcommand{\dalle}{\textsc{Dall$\cdot$e~2}\xspace}
\newcommand{\stablediffusion}{\textsc{Stable-Diffusion}\xspace}
\newcommand{\imagen}{\textsc{Imagen}\xspace}
\newcommand{\flamingo}{\textsc{Flamingo}\xspace}
\newcommand{\ingresnet}{\textsc{In-G-Rn}\xspace}
\newcommand{\ingvit}{\textsc{In-G-Vit}\xspace}
\newcommand{\lemon}{\textsc{Lemon}\xspace}
\newcommand{\clip}{\textsc{Clip}\xspace}
\newcommand{\lime}{\textsc{Lime}\xspace}
\newcommand{\gradcam}{\textsc{Grad-Cam}\xspace}
\newcommand{\imagenet}{\textsc{ImageNet}\xspace}
\newcommand{\imageneta}{\textsc{ImageNet-A}\xspace}
\newcommand{\imageneto}{\textsc{ImageNet-O}\xspace}
\newcommand{\imagenetc}{\textsc{ImageNet-C}\xspace}
\newcommand{\imagenetr}{\textsc{ImageNet-R}\xspace}
\newcommand{\imagenetnine}{\textsc{ImageNet-9}\xspace}
\newcommand{\waterbirds}{\textsc{Waterbirds}\xspace}
\newcommand{\wordnet}{\textsc{WordNet}\xspace}

\newcommand{\resnet}{\textsc{ResNet}\xspace}
\newcommand{\vit}{\textsc{Vit}\xspace}
\newcommand{\bit}{\textsc{Bit}\xspace}
\newcommand{\inception}{\textsc{Inception}\xspace}
\newcommand{\tfhub}{\textsc{Tf-Hub}\xspace}
\newcommand{\dittoclosing}{--- \raisebox{-0.5ex}{''} ---\xspace}

\newacronym{resnet}{\textsc{ResNet}}{Residual Network}
\newacronym{vit}{\textsc{Vit}}{Vision Transformer}

\theoremstyle{plain}

\theoremstyle{definition}

\theoremstyle{remark}

\usepackage[textsize=tiny]{todonotes}

\newcommand{\squishlist}{
   \begin{list}{$\bullet$}
    {\setlength{\itemsep}{0pt} \setlength{\parsep}{3pt}
     \setlength{\topsep}{3pt} \setlength{\partopsep}{0pt}
     \setlength{\leftmargin}{1em} \setlength{\labelwidth}{1em}
     \setlength{\labelsep}{0.5em}}}
\newcommand{\squishend}{
    \end{list}}

\usepackage[framemethod=tikz]{mdframed}
\mdfdefinestyle{mystyle}{%
  linecolor=black!80,
  fontcolor=black!80,
  leftline=true,
  innerleftmargin=10,
  innerrightmargin=0,
  outerlinewidth=3pt,
  topline=false,
  rightline=false,
  bottomline=false,
  skipabove=\topsep,
  skipbelow=\topsep
}

\usepackage{amsmath,amsfonts,bm}

\def\eqref#1{equation~\ref{#1}}

\def\1{\bm{1}}

\def\rvx{{\mathbf{x}}}

\def\rvz{{\mathbf{z}}}

\def\vs{{\bm{s}}}

\def\vx{{\bm{x}}}

\def\vz{{\bm{z}}}

\DeclareMathAlphabet{\mathsfit}{\encodingdefault}{\sfdefault}{m}{sl}
\SetMathAlphabet{\mathsfit}{bold}{\encodingdefault}{\sfdefault}{bx}{n}

\def\sA{{\mathbb{A}}}

\def\sD{{\mathbb{D}}}

\def\sS{{\mathbb{S}}}

\def\sX{{\mathbb{X}}}
\def\sY{{\mathbb{Y}}}
\def\sZ{{\mathbb{Z}}}

\newcommand{\E}{\mathbb{E}}

\newcommand{\lp}{\ensuremath{\ell_p}\xspace}

\DeclareMathOperator*{\argmax}{arg\,max}

\icmltitlerunning{Discovering Bugs in Vision Models using Off-the-shelf Image Generation and Captioning}

\begin{document}

\twocolumn[
\icmltitle{Discovering Bugs in Vision Models using Off-the-shelf \\ Image Generation and Captioning}

\icmlsetsymbol{equal}{*}

\begin{icmlauthorlist}
\icmlauthor{Olivia Wiles}{equal,yyy}
\icmlauthor{Isabela Albuquerque}{yyy}
\icmlauthor{Sven Gowal}{equal,yyy}
\end{icmlauthorlist}

\icmlaffiliation{yyy}{DeepMind, London, UK}

\icmlcorrespondingauthor{}{oawiles,sgowal,isabelaa@google.com}

\icmlkeywords{Machine Learning, ICML}

\vskip 0.3in
]

\printAffiliationsAndNotice{\icmlEqualContribution} %

\begin{abstract}
Discovering failures in vision classifiers under real-world settings remains an open challenge. This work shows how off-the-shelf, large-scale, image-to-text and text-to-image models, trained on vast amounts of data, can be leveraged to automatically find such failures. In essence, a conditional text-to-image generative model is used to generate large amounts of synthetic, yet realistic, inputs given a ground-truth label. A captioning model is used to describe misclassified inputs and descriptions are used in turn to generate more inputs, thereby assessing whether specific descriptions induce more failures than expected. As failures are grounded to natural language, we automatically obtain a high-level, human-interpretable explanation of each failure. We use this pipeline to interrogate classifiers trained on \imagenet to find specific failure cases and discover spurious correlations. Discovered failures generalize to other generative models and real images retrieved using {\em Google Image Search}. We also demonstrate the scalability of our approach by generating large adversarial datasets targeting specific classifier architectures. Finally, we discuss a number of further challenges. Overall, this work demonstrates the potential of large-scale generative models to automatically discover bugs in vision models in an open-ended manner. 
\end{abstract}

\section{Introduction}
Deep learning has enabled breakthroughs in a wide variety of fields~\citep{goodfellow_deep_2016, krizhevsky_imagenet_2012, hinton_deep_2012}, and deep neural networks are ubiquitous in many applications, including autonomous driving~\citep{bojarski_end_2016} and medical imaging~\citep{fauw_clinically_2018}.
Unfortunately, these models are known to exhibit numerous failures arising from using \emph{shortcuts} and \emph{spurious correlations} \citep{geirhos_shortcut_2020, arjovsky_invariant_2019, torralba_unbiased_2011, kuehlkamp_gender--iris_2017}.
As a result, they can fail when training and deployment data differ \citep{buolamwini_gender_2018}.
Hence, it is important to ensure that models are robust and generalize to new deployment settings.
Yet, only a few tools exist to automatically find failure cases on unseen data.
Some methods analyze the performance of models by collecting new datasets (usually by scraping the web).
These datasets must be large enough to obtain some indication of how models perform on a particular subset of inputs~\citep{hendrycks_natural_2019, hendrycks_many_2020, recht_imagenet_2019}.
Other methods rely on expertly crafted, synthetic (and often unrealistic) datasets that highlight particular shortcomings~\citep{geirhos_imagenettrained_2022, xiao_noise_2020}.

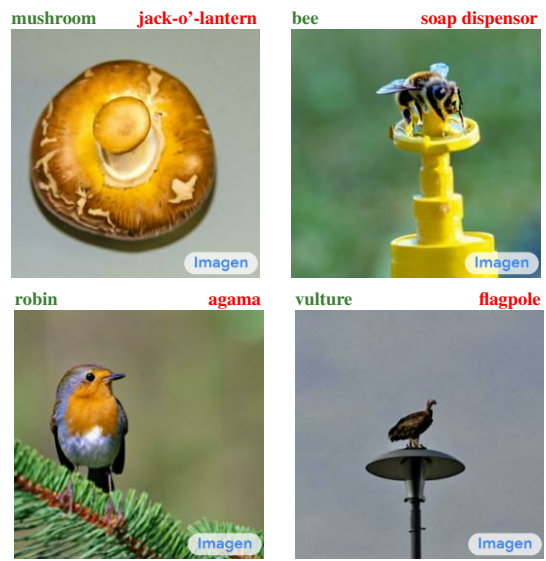
\begin{figure}[t]
    \centering
    \begin{overpic}[width=0.4\linewidth]{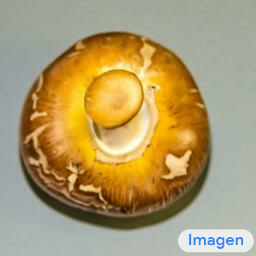}
    \put(0,102){\scriptsize \bf {\color{OliveGreen} mushroom}}
    \put(51,102){\scriptsize \bf {\color{red} jack-o'-lantern}}
    \end{overpic} \hspace{0.25cm}
    \begin{overpic}[width=0.4\linewidth]{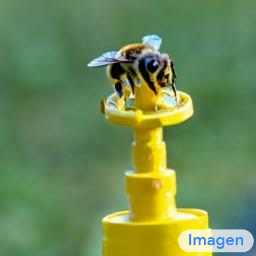}
    \put(0,102){\scriptsize \bf {\color{OliveGreen} bee}}
    \put(52,102){\scriptsize \bf {\color{red} soap dispensor}}
    \end{overpic} 
    \vspace{.4cm}
    
    \begin{overpic}[width=0.4\linewidth]{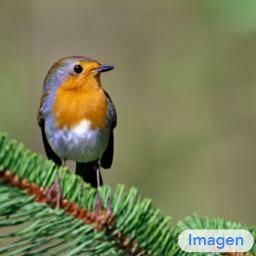} \hspace{0.5cm}
    \put(0,102){\scriptsize \bf {\color{OliveGreen} robin}}
    \put(78,102){\scriptsize \bf {\color{red} agama}}
    \end{overpic} \hspace{0.25cm}
    \begin{overpic}[width=0.4\linewidth]{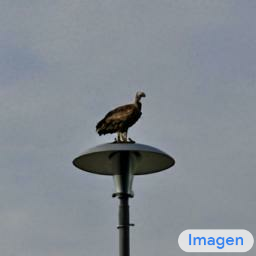}
    \put(0,102){\scriptsize \bf {\color{OliveGreen} vulture}}
    \put(74,102){\scriptsize \bf {\color{red} flagpole}}
    \end{overpic}
    \caption{Examples of automatically found failures. The correct label is to the left in {\color{OliveGreen} \bf green}. The incorrect prediction is to the right in {\color{red} \bf  red}. Images are not watermarked upon classification.\label{fig:ingteaser}}
\end{figure}

We present a methodology to automatically find failure cases (see \autoref{fig:ingteaser} for some examples) of image classifiers in an open-ended manner, without prior assumptions on the types of failures and how they arise.
We leverage off-the-shelf, large-scale, text-to-image, generative models, such as \dalle~\citep{ramesh_hierarchical_2022}, \imagen~\citep{saharia_photorealistic_2022} or \stablediffusion~\citep{rombach_highresolution_2022}, to obtain realistic images that can be reliably manipulated using a text prompt.
We leverage captioning models, such as \flamingo~\citep{alayrac_flamingo_2022} or \lemon~\citep{hu_scaling_2021}, to retrieve human-interpretable descriptions of each failure case.
This provides the following advantages:
\begin{enumerate*}[label={\it(\roman*)}]
\item generative models trained on web-scale datasets can be re-used and have broad non-domain-specific coverage; 
\item they demonstrate basic compositionality, can generate novel data and can faithfully capture the essence of (most) prompts, thereby allowing images to be realistically manipulated;
\item textual descriptions of these systematic failures that can be easily interpreted (even by non-experts) and interrogated (e.g., by performing counterfactual analyses).
\end{enumerate*}
Overall, our contributions are as follows:%

\begin{figure*}[t]
\centering
\resizebox{.9\textwidth}{!}{
\begin{tikzpicture}
\node[anchor=south west,inner sep=0] (image) at (0,0) {\includegraphics[width=1.0\textwidth]{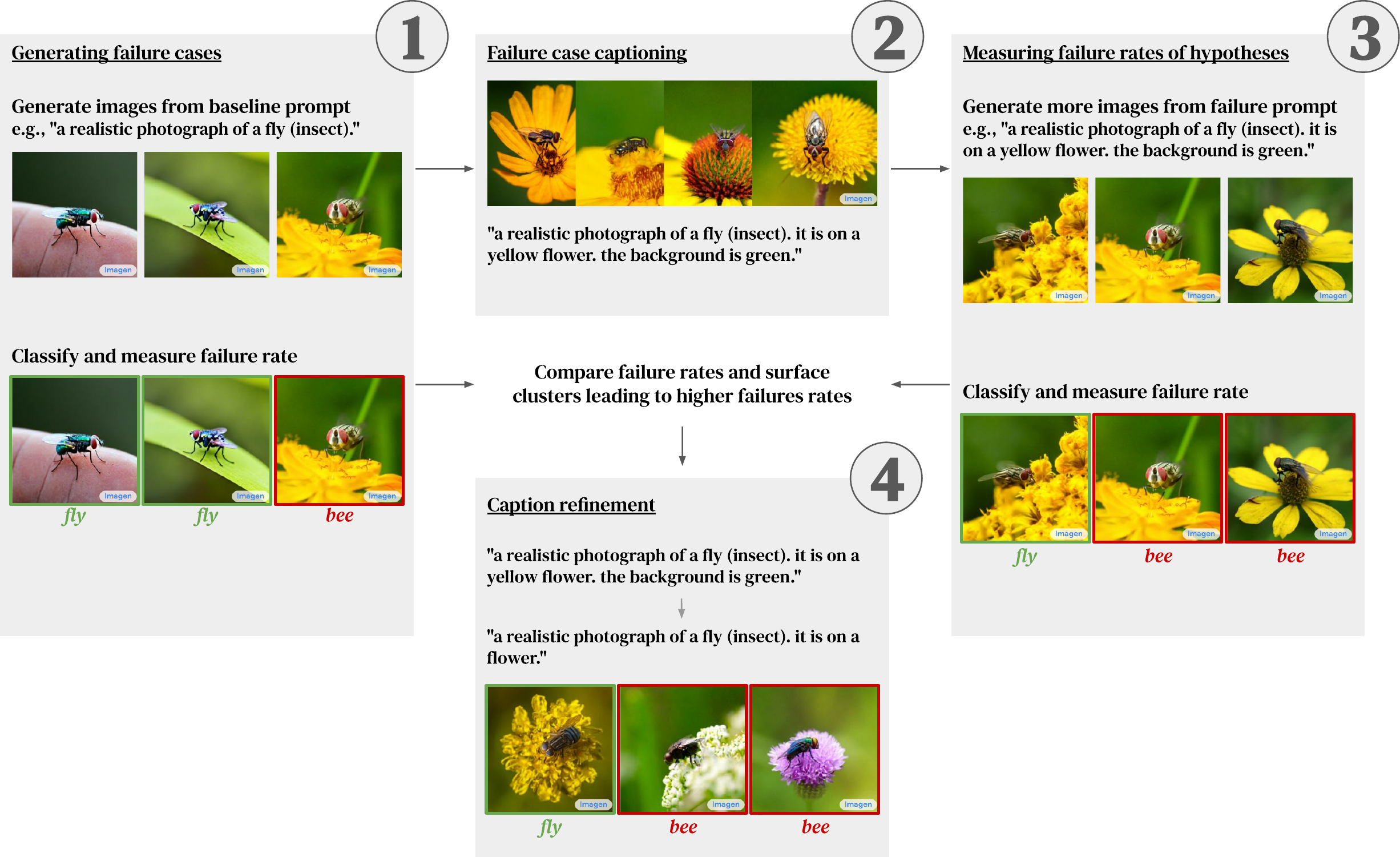}};
\node[black] at (2.5,6.7) {\tiny $\vx \sim \hat{p}(\rvx | y)$};
\node[black] at (2.5,3.7) {\tiny $\mathop{\E}_{\vx \sim p(\rvx|y)} \left [ f(\vx) \neq y \right ]$};
\node[black] at (8.2,7.0) {\tiny $\vz \sim \hat{p}(\rvz | \vx, y)$};
\node[black] at (14.1,6.3) {\tiny $\vx \sim \hat{p}(\rvx | \vz, y)$};
\node[black] at (14.1,3.1) {\tiny $\mathop{\E}_{\vx \sim p(\rvx|y,\vz)} \left [ f(\vx) \neq y \right ]$};

\end{tikzpicture}
}
\caption{\textbf{Diagram of our method.} The method starts by generating images containing a given class $y$ to measure the baseline failure rate of that class (right-hand side of \autoref{eq:1}). We construct a textual description for each misclassified image. This description is used to produce new images and measure the failure rate on images corresponding to that description (left-hand side of \autoref{eq:1}). The final description can be edited (manually or automatically) to understand the source of the failures.}
\label{fig:pipeline}
\end{figure*}
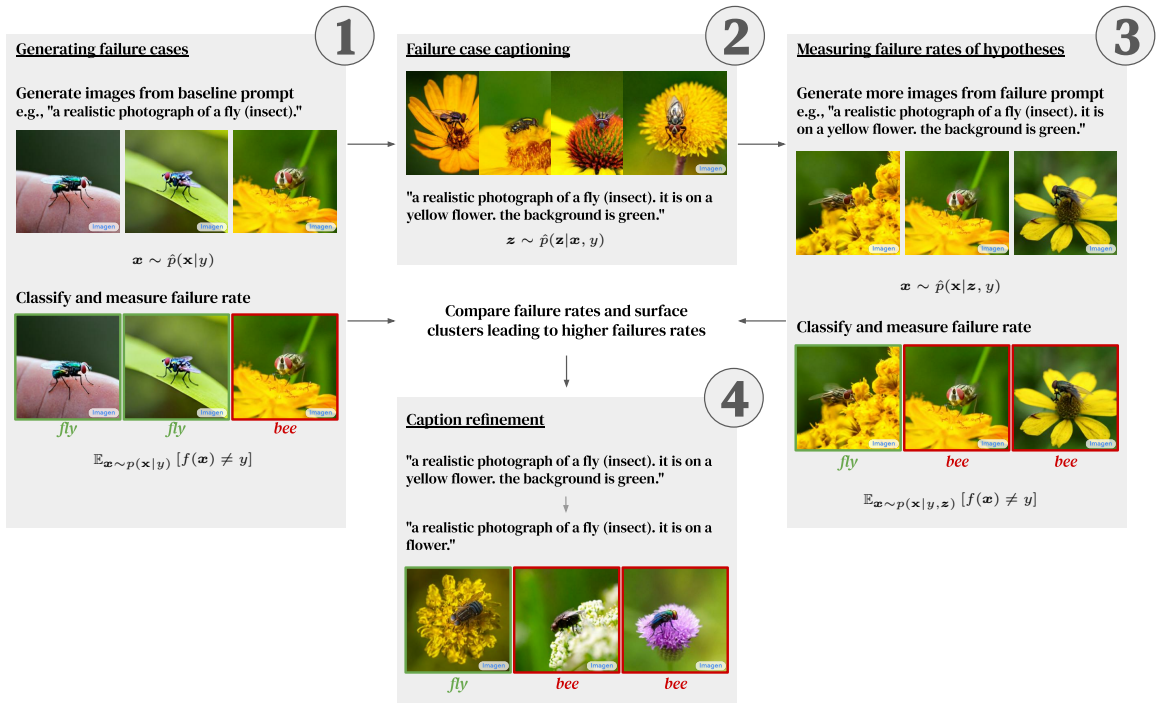

\squishlist
    \item We describe a methodology to discover failures of classifiers trained on \imagenet~\citep{deng_imagenet_2009}. To the contrary of prior work, we leverage off-the-shelf generative models, thereby avoiding the need to collect new datasets or rely on manually crafted synthetic images.
    \item Our approach surfaces failures that are human-interpretable by clustering and captioning inputs on which classifiers fail. These captions can be modified to produce alternative hypotheses of why failures occur allowing insights into the limitations of a given model. Furthermore, as failures are grounded to language, we demonstrate that they generalize across multiple text-to-image models and, more importantly, we show that similar failures occur with real images obtained via {\em Google Image Search}.
    \item We demonstrate the scalability of the approach by generating adversarial datasets (akin to \imageneta; \citealp{hendrycks_natural_2019}). In contrast to \imageneta, our new generated datasets align more closely with the original training distribution from \imagenet and generalize to multiple classifier architectures.
\squishend

Importantly, while this work focuses on vision models trained on \imagenet, it is neither limited to \imagenet nor the visual domain.
It serves as a proof-of-concept that demonstrates how large-scale, off-the-shelf, generative models~\citep{bommasani_opportunities_2021} can be combined to automate the discovery of \emph{bugs} in machine learning models and produce compelling, interpretable descriptions of model failures.
The approach is agnostic to the model architecture, which can be treated as a black box.

\vspace{-.2cm}
\section{Related Work}

\paragraph{Model failures.}

Spurious correlations can entice models to learn unintended shortcuts that obtain high accuracy on the training set but fail to generalize to new settings~\citep{lapuschkin_unmasking_2019, geirhos_shortcut_2020}.
\citet{recht_imagenet_2019} show that the accuracy of \imagenet models is impacted by changes in the data collection process, while \citet{torralba_unbiased_2011, khosla_undoing_2012, choi_context_2012} explore how contextual bias affects generalization.
\citet{mania_model_2019} demonstrate that models trained on \imagenet make consistent mistakes with one another and \citet{geirhos_beyond_2020} show that these mistakes are not necessarily consistent with human judgment.

\vspace{-.2cm}
\paragraph{Evaluation datasets.}

Understanding how model failures arise and empirically analyzing their consequences often requires collecting and annotating new test datasets.
\citet{hendrycks_natural_2019} collected datasets of natural adversarial examples (\imageneta and \imageneto) to evaluate how model performance degrades when inputs have limited spurious cues.
\citet{hendrycks_many_2020} collected four real-world datasets (including \imagenetr) to understand how models behave under distribution shifts.
In many cases, particular shortcomings can only be explored using synthetic datasets~\citep{cimpoi_describing_2013}.
\citet{hendrycks_benchmarking_2018} introduced \imagenetc, a synthetic set of common corruptions.
\citet{geirhos_imagenet-trained_2018} propose to use images with a texture-shape cue conflict to evaluate the propensity of models to over-emphasize texture cues.
\citet{xiao_noise_2020,sagawa2020distributionally} investigate whether models are biased towards background cues by compositing foreground objects with various background images (\imagenetnine, \waterbirds).
In all cases, building such datasets is time-consuming and requires expert knowledge.

\vspace{-.2cm}
\paragraph{Automated failure discovery.}

In some instances, it is possible to distill rules or specifications that constrain the input space enough to enable the automated discovery of failures via optimization or brute-force search.
In vision tasks, adversarial examples, which are constructed using \lp-norm bounded perturbations of the input, can cause neural networks to make incorrect predictions with high confidence \citep{carlini_adversarial_2017, carlini_towards_2017, goodfellow_explaining_2014, kurakin_adversarial_2016, szegedy_intriguing_2013}.
In language tasks, some efforts manually compose templates to generate test cases for specific failures \citep{jia_adversarial_2017, garg_counterfactual_2019, ribeiro_beyond_2020}.
Such approaches rely on human creativity and are intrinsically difficult to scale.
Several works \citep{baluja_adversarial_2017, song_constructing_2018, xiao_generating_2018, qiu_semanticadv:_2019, wong_learning_2021, laidlaw_perceptual_2020, gowal_achieving_2019} go beyond hard-coded rules by leveraging generative and perceptual models.
However, such approaches are difficult to automate as it is unclear how to relate specific latent variables to isolated structures of the original input.
Finally, we highlight a concurrent work \citep{ge2022dalle}, which leverages captioning and text-to-image models to construct background images to evaluate (and improve) an object detector.
Their approach requires compositing the resulting images with foreground objects and is not open-ended, in the sense that it requires a dataset of background images. 
Perhaps, the work by \citet{perez_red_2022} on \emph{red-teaming} language models is the most similar to ours.
\citeauthor{perez_red_2022} demonstrate how to prompt a language model to automatically generate test cases to probe another language model for toxic and other unintended output.

\vspace{-.2cm}
\paragraph{Interpretability.}

Interpretability techniques aim to give a rationale behind individual model predictions.
\citet{casper_2022} demonstrate that feature-level attacks, which create adversarial patches, can help diagnose brittle feature associations.
However, much like \lime~\citep{ribeiro2016why} or \gradcam~\citep{selvaraju2019gradcam}, the results are difficult to understand.
Other works \citep{abid2022meaningfully,jain2022distilling,eyuboglu2022domino} leverage auxiliary information in the form of attributes or image-to-text embeddings (e.g., from \clip; \citealp{radford_learning_2021}) to provide explanations in natural language.
However, these methods often rely on an additional dataset which limits their scope. 

\vspace{-.2cm}
\section{Method}
\label{sec:method}

\vspace{-.1cm}
\paragraph{Notation.}

We consider a classifier $f : \sX \to \sY$, where $\sX$ is the set of inputs (i.e., images) and $\sY$ is the label set.
We also assume that inputs $\vx \in \sX$ with label $y \in \sY$ are drawn from an underlying distribution $p(\rvx|\vz,y)$ conditioned on latent representations $\vz \in \sZ$.
In the context of this specific work, $\vz$ is a textual description of the image $\vx$.
We are interested in discovering captions $\vz$ corresponding to images $\vx \sim p(\rvx|\vz,y)$ with label $y$ that lead to significantly higher misclassification rates than generic images drawn from the marginal distribution $p(\rvx|y)$ conditioned solely on the label.
Formally, given a label $y$, we would like to find $\vz$ with
\begin{align}
    \mathop{\E}_{\vx \sim p(\rvx|\vz,y)} \left [ f(\vx) \neq y \right ] > \mathop{\E}_{\vx \sim p(\rvx|y)} \left [ f(\vx) \neq y \right ]
    \label{eq:1}
\end{align}
where $[\cdot]$ represents the Iverson bracket.
We may also be interested in identifying specific misclassifications towards a wrong (target) label $\bar{y} \neq y$, and would like that
\begin{align}
    \mathop{\E}_{\vx \sim p(\rvx|\vz,y)} \left [ f(\vx) = \bar{y} \right ] > \mathop{\E}_{\vx \sim p(\rvx|y)} \left [ f(\vx) = \bar{y} \right ] .
    \label{eq:2}
\end{align}

As we do not have access to the true underlying distributions $p(\rvx | \vz, y)$ and $p(\rvx | y)$, we leverage a large-scale text-to-image model, \imagen, to approximate them.
Similarly, we approximate $p(\rvz | \vx, y)$ with a captioning model, \flamingo.
We denote approximations of these distributions with the symbol $\hat{p}$.

\begin{mdframed}[style=mystyle]
For each of the following steps, we highlight additional implementation details and explain how we construct prompts for the text-to-image and image-to-text models.
\end{mdframed}

\vspace{-.2cm}
\paragraph{Generating failure cases.}

Our approach is described in \autoref{fig:pipeline}.
It consists of initially finding \emph{baseline} failures for the underlying model $f$ by sampling inputs $\vx$ from $\hat{p}(\rvx | y)$.\footnote{In practice, it is possible to steer the generation process to induce more frequent failures (e.g., by optimizing latents or conditioning via gradient ascent on the cross-entropy loss; ~\citealp{wong_learning_2021}). However, as a proof-of-concept, we assume only black-box access to off-the-shelf text-to-image model.}
Given a label of interest $y$, the output of this step is a set $\sD = \{ \vx_i \sim \hat{p}(\rvx | y) \}_{i=1}^N$ (where $N$ is the number of images we wish to generate), a set $\sD_\textrm{fail} = \{ \vx \in \sD | f(\vx) \neq y \}$ and an estimate of the baseline failure rate $|\sD_\textrm{fail}|/N$ (corresponding to the right-hand side of \autoref{eq:1}).
In this work, we consider problematic misclassifications only so we restrict ourselves to failures where any of the top-3 predicted labels are not under the same parent as the true label $y$ in the \wordnet hierarchy~\citep{miller_wordnet_1995}.\footnote{Other options include considering only misclassifications where the true label is not in the top-$k$ predicted labels or where the confidence in the true label is lower than a predefined threshold.}

\begin{mdframed}[style=mystyle]
This step leverages a text-to-image generative model conditioned on $y$ with prompts such as \textit{``a realistic photograph of a fly (insect).''}, which are automatically generated from the corresponding class and \wordnet hierarchy.
As our domain of interest is composed of real images, the prompt is designed to enforce the generation of photographs of real objects and animals rather than drawings or paintings.
Further prompt engineering can be explored to find failures on a  variety of domains (such as sketches or medical imaging; \citealp{hendrycks_many_2020,kather_medical_2022}).
\end{mdframed}

\paragraph{Optional clustering failure cases.}

Clustering is not necessary, but makes the search for the caption $\vz$ leading to high failure rates more efficient and computationally manageable.
First, we split $\sD_\textrm{fail}$ by predicted label, i.e. $\sD_\textrm{fail} = \smash{\sD_\textrm{fail}^{(1)}} \cup \ldots \cup \smash{\sD_\textrm{fail}^{(|\mathcal{Y}|)}}$, where $\smash{\sD_\textrm{fail}^{(y)}} = \{ \vx \in \sD_\textrm{fail} | f(\vx) = y \}$.
Then, for each subset $\smash{\sD_\textrm{fail}^{(y)}}$, we group inputs that have similar feature representations together (e.g., using the cosine distance between intermediate activations of a pretrained model).
The goal of this step is to reduce the number of clusters to a minimum without amalgamating different causes of failure together.\footnote{Wrongful clustering may lead the next step to produce descriptions that fail to induce more failures. As a result, we may miss failures that we would have discovered without clustering, but the failures that we do discover remain valid.}

\begin{figure*}[t]
\centering
\subfigure[ \scriptsize \texttt{Persian cat} $\rightarrow$ \texttt{Snow leopard}\label{fig:openended_failures_cat}]{\includegraphics[width=0.32\textwidth]{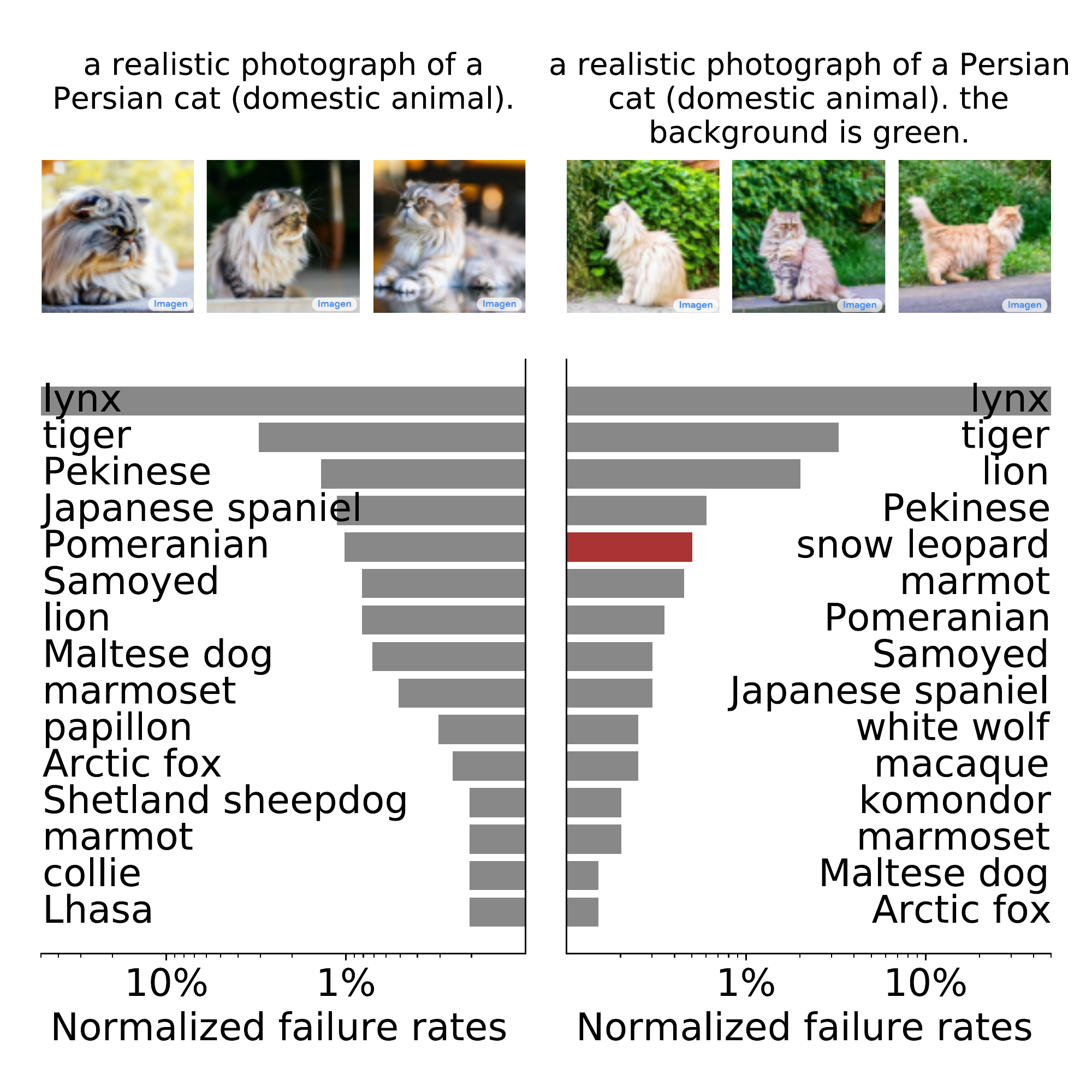}}
\subfigure[ \scriptsize \texttt{Fly} $\rightarrow$ \texttt{Bee}\label{fig:openended_failures_fly}]{\includegraphics[width=0.32\textwidth]{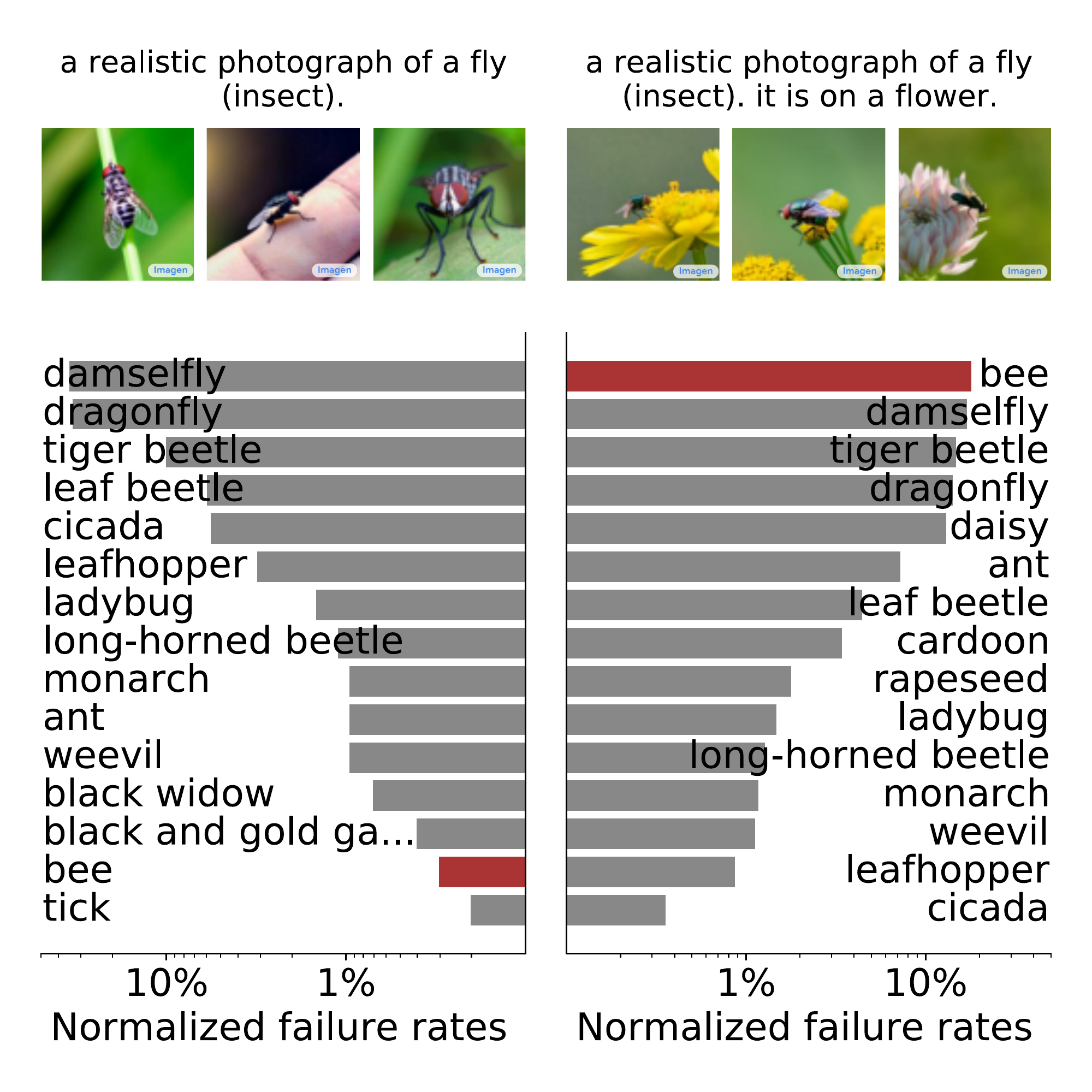}}
\subfigure[ \scriptsize \texttt{Crayfish} $\rightarrow$ \texttt{Chainlink fence}\label{fig:openended_failures_crayfish}]{\includegraphics[width=0.32\textwidth]{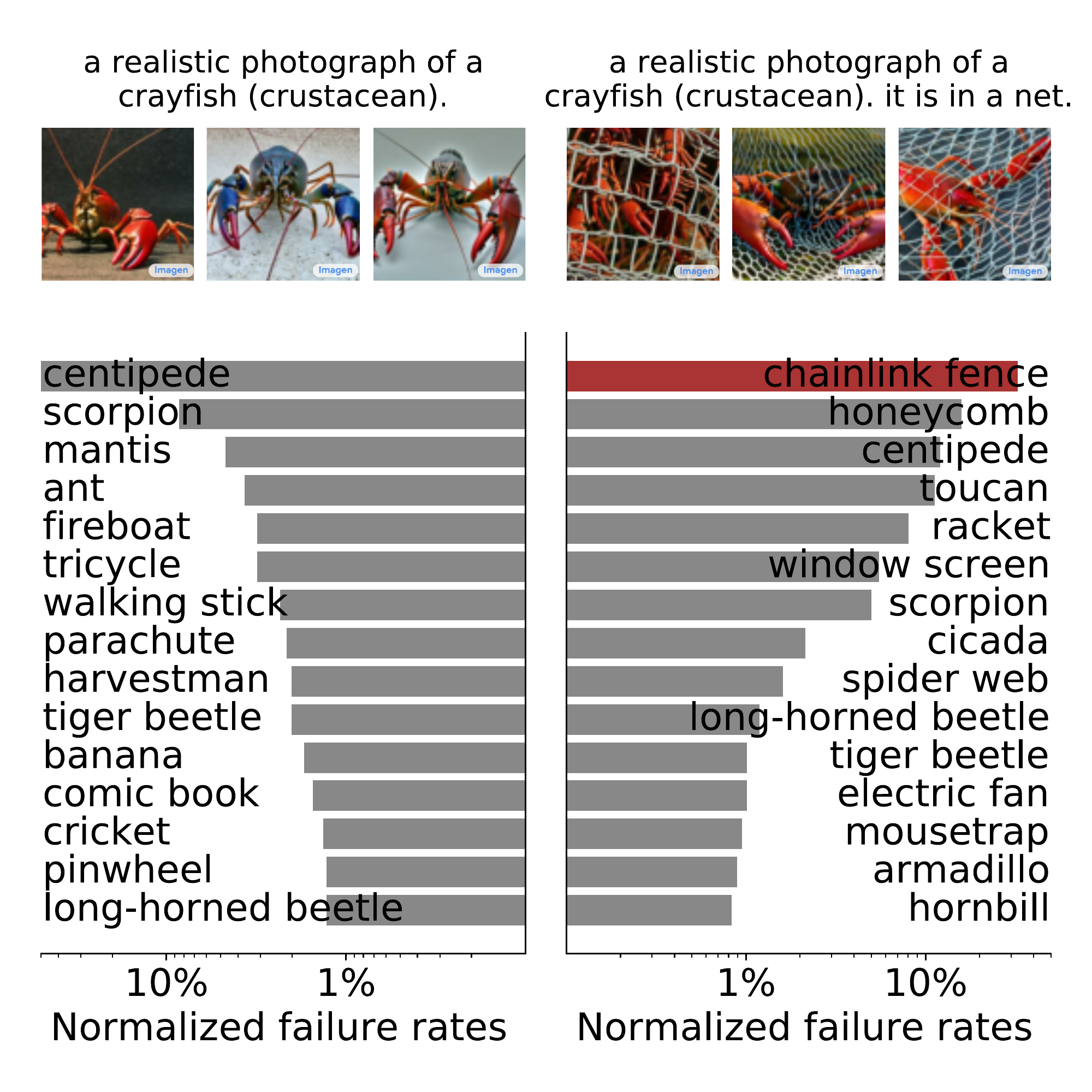}}
\caption{\textbf{Distribution of failures of a \resnet-50 for the baseline and automatically discovered captions for three true and target label pairs.} For each panel, failures resulting from the baseline caption are on the left and failures resulting from the discovered caption are on the right. We show the top-15 mistakes and three randomly sampled images for each caption. We highlight in red the bar corresponding to the target label. Absolute failure rates are given in \autoref{table:openended_failures}.}
\label{fig:openended_failures}
\end{figure*}

\paragraph{Failure case captioning.}

For each cluster $\sA \subseteq \sD_\textrm{fail}$, we would like to find a caption $\vz_\sA$ that describes it. Grounding failures in simple textual descriptions allows us to maintain the diversity of the generated images: the generated images resemble the images leading to the original failure without being exact copies.
Ideally, we would like to find the caption $\vz_\sA$ that maximizes the likelihood of sampling elements of $\sA$, i.e., $\vz_\sA = \smash{\argmax_\vz \prod_{\vx \in \sA} \hat{P}(\vx | \vz, y)}$.
We may wish to impose constraints on $\vz_\sA$, such as a maximum number of words or sentences.\footnote{These constraints guarantee that captions remain simple and not overly descriptive.}
Finding such a caption is computationally hard and measuring exact likelihoods $\hat{P}(\vx | \vz, y)$ can be challenging.
Hence, we resort to sampling captions directly from a captioning model $\hat{p}(\rvz | \vx, y)$ for each image of cluster $\sA$.\footnote{This formulation implicitly assumes that any caption is as likely as another under a given label $y$, which in general does not hold true, but serves as a reasonable approximation.}
Captions are split into sentences, resulting in a set of sentences $\sS$.
Sentences are greedily combined (up to a maximum number of sentences $K$) to maximize the likelihood of sampling the overall caption $\vz_\sA = [\vs_1, \ldots, \vs_K]$ with $\vs_j = \smash{\argmax_{\vs \in \sS} \prod_{\vx \in \sA} \hat{P}([\vs_1, \ldots, \vs_{j-1}, \vs] | \vx, y)}$ and $\vs_1 = \smash{\argmax_{\vs \in \sS} \prod_{\vx \in \sA} \hat{P}([\vs] | \vx, y)}$.
If no clustering is performed, we can directly sample a caption $\smash{\vz_{\{ \vx \}}} \sim \hat{p}(\rvz | \vx, y)$ from the captioning model for each element $\vx$ in $\sD_\textrm{fail}$.
Each caption serves as a failure hypothesis.

\begin{mdframed}[style=mystyle]
This step leverages an image-to-text captioning model and forces descriptive completions via few-shot prompting.
In particular, we use three shots, with each shot being an image and caption pair.
Images are publicly available online and are manually described using between four to seven short sentences that describe the subject of the photograph and its physical position with respect to the camera, as well as the background or context in which the subject appears (see \autoref{sec:prompt} for more details).
To condition on $y$, we force the captioning model to only consider completions to the original baseline prompt, which guarantees that the final caption contains the true label.
An example of a resulting caption is \textit{``a realistic photograph of a fly (insect). the background is blurred. the fly is in focus. it is on a yellow flower. the background is green.''}
Each caption serves as a failure hypothesis.
\end{mdframed}

\paragraph{Measuring failure rates of hypotheses.}

For each failure hypothesis or caption $\vz_\sA$, we can measure its failure rate via sampling $\mathop{\E}_{\vx \sim p(\rvx|\vz_\sA,y)} \left [ f(\vx) \neq y \right ]$.
This step allows us to surface captions $\vz^\star$ that satisfy \autoref{eq:1} (or alternatively \autoref{eq:2}) by comparing the resulting failure rate with the baseline failure rate obtained initially.

\begin{mdframed}[style=mystyle]
This step uses the text-to-image model, which we now prompt with descriptive cluster captions, e.g.~\textit{``a realistic photograph of a fly (insect). the background is blurred. the fly is in focus. it is on a yellow flower. the background is green.''}
\end{mdframed}

\paragraph{Caption refinement and counterfactual analysis.}

Given a caption $\vz^\star$, we want to provide a shorter, self-contained caption that obtains a similar failure rate.
For this step, we rely on simple rules\footnote{This may seem at odds with our claim on open-endedness. However, we note that this step is optional and its goal is simply to produce shorter failure descriptions.} and evaluate promising caption rewrites.
Finally, as captions are human-readable, users can interact with the system and test alternative hypotheses.

\begin{mdframed}[style=mystyle]
In our implementation, we exploit two rules:
\begin{enumerate*}[label={\it(\roman*)}]
\item we evaluate all individual sentences (for our ongoing example, these would be \textit{``the background is blurred''},  \textit{``the fly is in focus''}, \textit{``it is on a yellow flower''} and \textit{``the background is green''}) in conjunction with the original prompt (e.g., \textit{``a realistic photograph of a fly (insect)''});
\item the most promising prompt is further refined by dropping adjectives (such as \textit{``yellow''} in \textit{``it is on a yellow flower''}).
\end{enumerate*}
More sophisticated rules and rewrites are possible \citep{ribeiro_semantically_2018}.
Alternatively, we could leverage a large-language model via few-shot prompting to automate this process \citep{witteveen_paraphrasing_2019}.
\end{mdframed}

Finally, we note that it is possible to use part of this pipeline to understand known failure cases (e.g., on failures reported by external users of the model $f$).

\section{Results}

We elaborate on two use-cases.
First, we find failure cases of a \gls{resnet}-50~\citep{he_deep_2015} trained on \imagenet.
We focus on three arbitrarily chosen labels and show that the failures we obtain arise from consistent misclassifications caused by spurious correlations.
We show that these failures generalise to {\em real} images downloaded through {\em Google Image Search}.
Second, we generate failures at scale for various models and show that these failures generalise to other architectures and model initializations.

\vspace{-.2cm}
\subsection{Open-ended failure search}

\begin{table*}[t]
\begin{center}
\resizebox{\textwidth}{!}{
\sisetup{input-symbols=\%}
\begin{tabular}{ll|p{9cm}|lr|lr}
    True label & Target label & Caption & \multicolumn{2}{l|}{Failure rate (any)} & \multicolumn{2}{l}{Failure rate (target)} \\
    \toprule
    \multirow{2}{*}{\texttt{Persian cat}} & \multirow{2}{*}{\texttt{Snow leopard}} & a realistic photograph of a Persian cat (domestic animal). & ~~0.11\% & 1$\times$ & ~~0.00022\% & 1$\times$ \\
    & & \dittoclosing the background is green. & ~~0.64\% & 6$\times$ & ~~0.0032\% & 14$\times$ \\
    \hline
    \multirow{2}{*}{\texttt{Fly}} & \multirow{2}{*}{\texttt{Bee}} & a realistic photograph of a fly (insect).& ~~0.48\% & 1$\times$ & ~~0.0014\% & 1$\times$ \\
    & & \dittoclosing it is on a flower. & ~~4.11\% & 9$\times$ & ~~0.72\% & 497$\times$ \\
    \hline
    \multirow{2}{*}{\texttt{Crayfish}} & \multirow{2}{*}{\texttt{Chainlink fence}} & a realistic photograph of a crayfish (crustacean). & ~~0.93\% & 1$\times$ & ~~0.00047\% & 1$\times$ \\
    & & \dittoclosing it is in a net. & ~~6.31\% & 7$\times$ & ~~1.73\% & 3721$\times$ \\
    \bottomrule
\end{tabular}
}
\end{center}
\caption{\textbf{Absolute failure rates of a \resnet-50 for three true and target label pairs.} We show the total failure rate (i.e., the model prediction is different from the true label) and the target failure rate (i.e., the model prediction is the target label). Captions are automatically discovered using the method detailed in \autoref{sec:method}.\protect\footnotemark\label{table:openended_failures}}
\end{table*}

\vspace{-.2cm}
\paragraph{Setup.}

We evaluate a \resnet-50 trained on \imagenet and available on \tfhub.\footnote{\url{https://tfhub.dev/google/imagenet/resnet_v2_50/classification/5}}
We select three labels $y$ at random: \texttt{Persian cat}, \texttt{fly} and \texttt{crayfish}.
For each label $y$, we manually select target labels $\bar{y}$ (\texttt{snow leopard}, \texttt{bee} and \texttt{chainlink fence} respectively) and execute the protocol defined in \autoref{sec:method}.
At each step, we sample images from the generative model until we gather 20 images that are misclassified as the target label $\bar{y}$ and compute failure rates at that point.
\autoref{fig:openended_failures} and \autoref{table:openended_failures} show these automatically discovered failures.
More failures for additional true and target label pairs are in \autoref{sec:additional_openended} in the appendix.
In \autoref{sec:additional_openended}, we also evaluate other architectures (i.e., \vit{s}) and demonstrate that the discovered captions yield failure rates that are statistically significant.

\paragraph{Discovered failures.}

\autoref{fig:openended_failures_cat} shows the distribution of failures for the baseline label \texttt{Persian cat}.
We observe that the most frequent confusion, on images generated using the baseline caption \textit{``a realistic photograph of a Persian cat (domestic animal).''} is with \texttt{lynx}.
This mistake arises about 0.1\% of the time and constitutes 87.3\% of all failures.
In comparison, the confusion with \texttt{snow leopard} is rather infrequent and arises only 0.00022\% of the time.
Our approach automatically discovers that adding \textit{``the background is green.''} to the caption results in a large increase in failure rates.
Failures are 5.72$\times$ more likely and the model is 14.3$\times$ more likely to predict \texttt{snow leopard}.
We generally observe that mistakes with wild animals become more prevalent when the cat is outdoors.
Similarly, \autoref{fig:openended_failures_fly} and \autoref{fig:openended_failures_crayfish} show failures on images of flies and crayfish, respectively.
Flies on flowers are significantly more likely to be confused for bees when they are on flowers (497$\times$), while crayfish in nets are more frequently confused as chainlink fences (3721$\times$), honeycomb, window screens or spider webs.
These highlight two shortcomings of the underlying classifier:
\begin{enumerate*}[label={\it(\roman*)}]
\item the over-reliance on spurious cues (such as the flower);
\item the inability to determine which object is the main subject of a photograph (e.g., which of the net or crayfish is important).
\end{enumerate*}

\footnotetext{As a point of comparison, we can also evaluate the baseline failure rates on images from the \imagenet test set. For \texttt{Persian cat}, \texttt{fly} and \texttt{crayfish} the baseline failure rates are 16\%, 8\% and 18\%, respectively (the target failure rate is 0\% for all labels). These failure rates are higher than on images generated by the generative model, this is perhaps indicating that the generative model produces images that are more canonical and conservative. We also point out that we generate 9.1M samples to compute the first row of the table, 625K for the second and 1.4M, 2.8K, 4.3M, 1.2K for the subsequent rows.}

\paragraph{Generalization of failure descriptions.}

To verify that the discovered failures are not specific to the text-to-image model used in this manuscript and do not result from artifacts in the image generation process, we generate 30 images using the baseline and discovered captions with \dalle and \stablediffusion (samples are shown in \autoref{fig:dalle} and \autoref{fig:stablediffusion} in the appendix).
We evaluate the failure rates for the \texttt{fly} and \texttt{crayfish} labels (which exhibited higher failure rates).
With \dalle, for the 30 images generated with the prompt\textit{ ``a realistic photograph of a fly (insect).''}, 18 are correctly classified as flies and none  as bees.
When adding \textit{``it is on a flower.''} to the prompt, the overall failure rate increases (only 14 images are correctly classified) and nine images are now classified as bees.
Similarly, for \textit{``a realistic photograph of a crayfish (crustacean).''}, 29 images are correctly classified as \texttt{crayfish}, \texttt{spiny lobster}, \texttt{American lobster}, \texttt{Dungeness crab} or \texttt{king crab}, while none are classified as \texttt{chainlink fence}.
When adding \textit{``it is in a net.''}, four are classified as \texttt{chainlink fence} (with \texttt{chainlink fence} appearing ten times in the top-3 predictions), while only 21 images are correctly classified.
Results are similar for \stablediffusion images.\footnote{The number of failures increases from one to two and from four to 16, for the \texttt{fly} and \texttt{crayfish} labels respectively.}
Overall, we observe that discovered failures generalize across generative models.

\paragraph{Generalization to Google Image Search.} Finally, we verify that failures generalise to images queried through \emph{Google Image Search}\footnote{\url{https://images.google.com}}. We query \emph{Google Image Search} to find 30 images for each of the following queries: \textit{``fly''}, \textit{``fly on flower''}, \textit{``crayfish''}, \textit{``crayfish in net''} (images must have a resolution of at least 256$\times$256 and should contain the true label).
We classify all images and observe that the number of failures towards \texttt{bee} increases from zero to two and those towards \texttt{chainlink fence} increase from zero to four.
This illustrates that the discovered failures are general and even extend to real photographs.
Further examples are given in \autoref{sec:additional_openended}.

\subsection{Adversarial dataset generation at scale}

We demonstrate how to apply our automated pipeline to generate large datasets of failures.
We seed our search by captioning images from \imageneta.
We show that the discovered failures generalize across initializations of a given model architecture and between models of different architectures.

\paragraph{Generating large-scale datasets.}

We assume that we have access to a set of captions that describe potential failure cases.\footnote{This assumption is not necessary. However, it accelerates our search by generating images that are more likely to induce failures.}
These captions are automatically extracted from the 7,500 images of \imageneta using the captioning model, limiting its output to two sentences maximum.
For each caption, we sample up to a thousand images keeping those leading to misclassifications and limit the number of misclassified images kept per caption.
We consider two models: a \gls{resnet}-50 (abbreviated hereafter by \textsc{Rn}) and a Vision Transformer in its B/16 variant \citep{dosovitskiy_image_2020}, abbreviated by \vit.
Both models are trained solely on \imagenet and achieve 76\% and 80\% top-1 accuracy, respectively. 
This yields two separate datasets of failures which we refer to as \ingresnet and \ingvit that are of size 12,332 and 9,536 respectively.

\begin{figure}[t]
    \includegraphics[width=\textwidth,trim={0cm 0.6cm 0cm 0cm},clip]{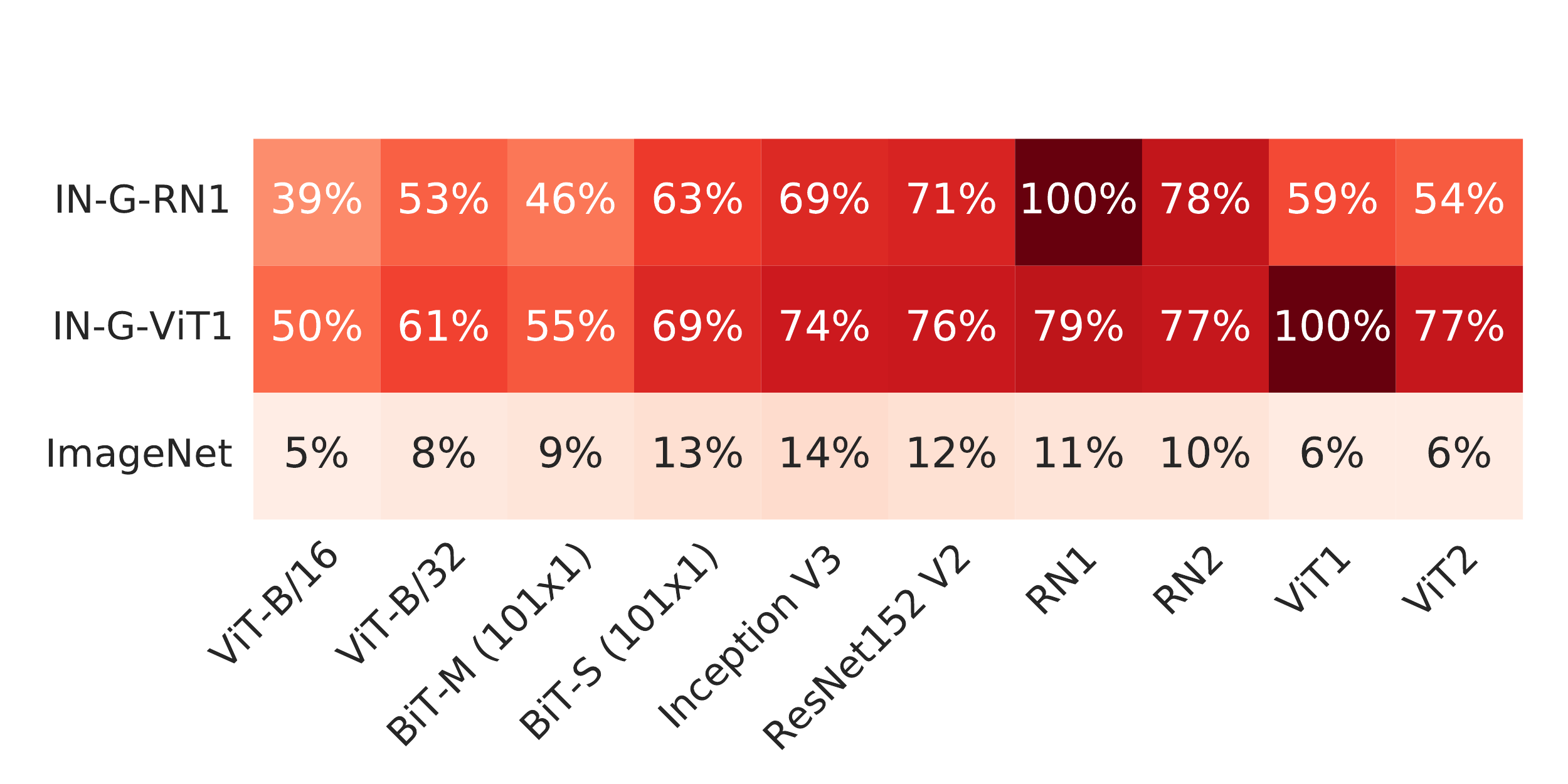}
    \caption{{\bf Failure rates (top-3) for different models on two generated datasets and \imagenet.} We report the failure rates of different models trained on \imagenet\label{fig:modelcomparisons}.}
\end{figure}

\paragraph{Visualizations of failure cases.}

\autoref{fig:ingteaser} shows samples from \ingresnet with true and predicted labels.
While the images clearly show the correct class, the model erroneously predicts a different one.
Additional samples for \ingresnet and \ingvit are in \autoref{fig:ingresnet} and \autoref{fig:ingvit} (in the appendix).
\autoref{sec:interpreting} provides an analysis for some failure cases found in \ingvit and details how images from the \imagenet training set may have misled the classifier.

\paragraph{Generalization to other networks.}
We verify that our failures generalize to new architectures and initializations of a specific model to demonstrate that the failures are general and robust.
We train an additional \gls{resnet} (\textsc{Rn2}) and \vit (\textsc{Vit2}) on \imagenet with the same setup as our two original models but different random seeds.
We also consider a large set of additional models trained on \imagenet and optionally pre-trained on larger datasets obtained from \tfhub.
\autoref{fig:modelcomparisons} shows the failure rates induced by both datasets on all models (failures are accounted when the top-3 predictions do not include the true label).
First, we observe that failures transfer well between models of the same architecture.
Indeed, 78\% of the failures in \ingresnet transfer to \textsc{Rn2}, while the ones in \ingvit transfer with 77\% chance to \textsc{Vit2}.
Second, we observe that failures for a given model architecture transfer to a large extent across architectures.
Even when large-scale pretraining is used (with the \bit{-M (101x1)}, \vit{-B/16} and \vit{-B/32} models pretrained on \imagenet{21K}), failures transfer at a rate of 39-53\% for \ingresnet and 50-61\% for \ingvit.
Further results for additional models are in \autoref{app:transferstudy} in the appendix.

\paragraph{Distribution shift.}
\label{sec:distributionshift}
We compare our generated datasets to \imagenet and \imageneta to validate that we generate images that are similar to those in \imagenet. 
We compute the Fr\'echet Inception Distance (FID; \citealp{heusel_gans_2017}) and the Kernel Inception Distance (KID; \citealp{binkowski2018demystifying}) between the generated images and the \imagenet test set.
\autoref{tab:distribution_shift} also shows the FID and KID of the \imagenet train set and \imageneta.
We find that our generated images are {\em more} similar to those from \imagenet under both metrics than those from \imageneta.

\begin{table}[t]
\centering
\footnotesize
\begin{tabular}{lcc}
     Dataset & FID $\downarrow$ & KID $\downarrow$ \\ \toprule
     \imageneta & 56.6 & 0.0460 \\
     \ingresnet & 48.3 & 0.0305 \\
     \ingvit & 53.9 & 0.0330 \\ \midrule
     \imagenet (train) & 2.3 & 0.0003 \\ \bottomrule
\end{tabular}
\caption{{\bf FID and KID scores.} We report FID and KID scores in relation to \imagenet (test).\label{tab:distribution_shift}}
\end{table}

\begin{figure*}[t]
\centering
\subfigure[Coverage \label{fig:limitations_a}]{\includegraphics[width=0.19\textwidth]{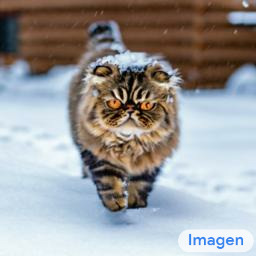}}
\subfigure[Bias \label{fig:limitations_b}]{\includegraphics[width=0.19\textwidth]{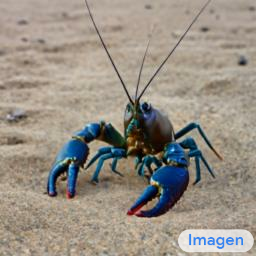}}
\subfigure[Captioning issues \label{fig:limitations_c}]{\includegraphics[width=0.19\textwidth]{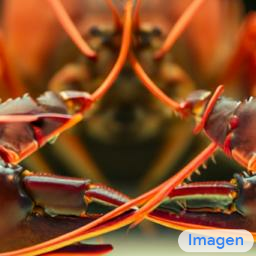}}
\subfigure[Generation issues \label{fig:limitations_d}]{\includegraphics[width=0.19\textwidth]{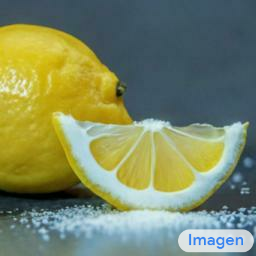}}
\subfigure[OOD sampling \label{fig:limitations_e}]{\includegraphics[width=0.19\textwidth]{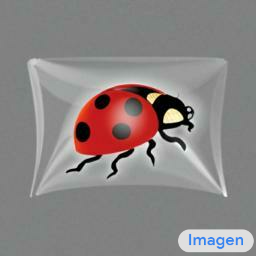}}
\caption{\textbf{Illustrative examples of various challenges.}
\subref{fig:limitations_a} Persian cats in snow (generated using \textit{``a realistic photograph of a Persian cat (domestic animal). it is walking in the snow.''}) are misclassified as snow leopards at a rate of 0.016\%, which is significantly higher than the failure rate of 0.0032\% induced by the automatically found caption (\textit{``\dittoclosing the background is green.''}); the total failure rate also increases twelve-fold to 8.15\% (from 0.64\%).
\subref{fig:limitations_b} It is estimated that only 1 in 10,000 crayfish turn blue. However, 9\% of the images generated using \textit{``a realistic photograph of a crayfish (crustacean).''} contain a blue crayfish (estimated by manually looking at 100 samples).
\subref{fig:limitations_c} This image of a crayfish is misclassified as a chainlink fence. The output of the captioning model for this particular image is \textit{``a realistic photograph of a crayfish. the crayfish is very detailed. the crayfish is facing the camera. the crayfish is orange. it has two antennae.''} While the caption describes the image, it does not provide enough detail to reconstruct the image.
\subref{fig:limitations_d} This image is generated from the caption \textit{``a realistic photograph of a saltshaker (container). there is a lemon slice on the side of the salt shaker.''} While the image contains a lemon, the true class $y$ (\texttt{saltshaker}) is not visible.
\subref{fig:limitations_e} Generated with the caption \textit{``a realistic photograph of a ladybug (insect). it is in a plastic bag.''}, this image illustrates that text-to-image models can create image that are not from the intended distribution (i.e., of realistic photographs).
}
\label{fig:limitations}
\end{figure*}

\section{Discussion}\label{sec:discussion}

Our work demonstrates that today's large-scale text-to-image and image-to-text models can be leveraged to find human-interpretable failures in vision models.
We have demonstrated these failures generalize to real images, other architectures, and can be obtained at scale.
While we focus on \imagenet, there are encouraging signs that generative models could be used to probe models trained on specialized tasks such as medical imaging~\citep{kather_medical_2022}.
Overall, there remain a number of key challenges to address.

\vspace{-.2cm}
\paragraph{Coverage.}

While our approach can be used to demonstrate the presence of failures, it is important to note that (just like scraping the web or using a fixed dataset) it cannot prove their absence: there is no guarantee that it will discover all failures of a given model.\footnote{If the generative model matches the true distribution of images, it is possible to give meaningful probabilistic guarantees.}
Moreover, the generative model is only an approximation of the distribution of interest and may lack coverage.
For example, it might almost never generate ``a lawnmower falling down from the sky'' (an actual image from the \imagenet training set; \citealp{jain2022databased}) when prompted with \textit{``a realistic photograph of a lawnmower''}.
While this can help ground failures to scenes that are likely to occur in the real-world, it also means that rare failures are unlikely to be discovered (see \autoref{fig:limitations_a}).

\vspace{-.2cm}
\paragraph{Bias.}

While we take the view that off-the-shelf large-scale generative models are trained on diverse and unbiased data, these models mirror the distribution of images and captions seen on the web.
The generative model may over-sample particular regions of the image manifold and, as a result, our approach is more likely to discover failures in these high-density regions while missing failures pertaining to other regions (see \autoref{fig:limitations_b}).
(This is similar to how \imagenet over-represents dogs.)
Possible solutions to reduce bias include clever prompting (which introduces expert knowledge) or discovering failure prompts more actively by avoiding random sampling (e.g., through adversarial techniques).

\vspace{-.2cm}
\paragraph{Captioning issues.}

Using captions allows our approach to produce human-interpretable explanations and constrains our search to failures that can be explained in words.
Not only is it possible for the captioning model to miss important details or produce ungrounded captions, but some failures may simply be hard to describe (even by a human).
As a result, newly generated images may look different from the set of images that induced the original failure.
We note that efficiently enforcing consistency between the generated and original images (through a common caption) is an open problem since we would like to search over \emph{reasonable} captions that are likely to produce images corresponding to the original failure.
\autoref{fig:limitations_c} shows an example where a \texttt{crayfish} is misclassified as a \texttt{chainlink fence}.
While the reason for that failure is immediately obvious to us, it remains difficult to describe with a succinct caption.

\vspace{-.2cm}
\paragraph{Out-of-distribution sampling.}

Ensuring that images sampled from an off-the-shelf generative model are part of the intended distribution (e.g., resembling \imagenet) is difficult.
We start our prompts with \textit{``a realistic photograph''} in a bid to help steer the approximated distribution $\hat{p}(\rvx | y, \vz)$ away from artistic drawings and closer to the true distribution $p(\rvx | y, \vz)$.
This approach is effective (as our images are statistically more similar to \imagenet than those in \imageneta are as discussed in \autoref{sec:distributionshift}), but not always successful (see \autoref{fig:limitations_e}).
In some cases, finding a suitable prompt is not obvious (e.g., to output images from a particular medical domain; \citealp{kather_medical_2022}) and fine-tuning models on the dataset of interest may be necessary.

\vspace{-.2cm}
\paragraph{Image generation issues.}

While the text-to-image model may make occasional mistakes (e.g.~generating the wrong object 1.45\% of the time for unambiguous prompts -- see \autoref{app:imagenetaccuracy}), subtle errors can arise.
The prompt can be ambiguous due to the presence of homonyms (e.g., a \textit{``walking stick''} can be both a cane or an insect), or may describe multiple objects with complex relationships that exacerbate mistakes (see \autoref{fig:limitations_d}).

\vspace{-.2cm}
\paragraph{Privacy.}

As we are generating large amounts of data, it is important to consider the associated privacy risks.
While these risks can be mitigated by using generative models trained on public, non-sensitive data, more research on private generative modelling is necessary \citep{harder_differentially_2022}.

Despite these challenges, we foresee that large-scale generative models will increasingly be used as debugging tools.
Moreover, as large-scale models improve so will our pipeline.
In this work, we introduced an automated pipeline that discovers failure cases in vision models. 
Our work demonstrates that such a system allows for large-scale investigations of vision models in an open-ended manner as it is automatic and failures are human-intepretable and generalize to other architectures and real images.

\bibliography{bibliography}
\bibliographystyle{icml2023}

\newpage
\appendix
\onecolumn

\section{Prompting the image-to-text model} \label{sec:prompt}

To ensure that captions are descriptive and composed of short sentences, we prompt our image-to-text model with the following for all experiments:

\begin{tabular}{c m{10.5cm}}
    \begin{minipage}{3.5cm}\includegraphics[height=3cm]{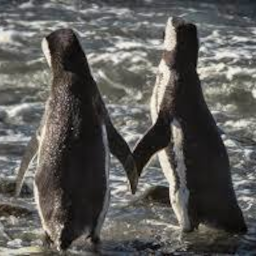}\end{minipage}
    & is a realistic picture of two penguins. They are holding hands. They are standing in front of the sea. The picture is mostly grey. The penguins are facing away from the camera. They take up most of the image. \\
    \begin{minipage}{3.5cm}\includegraphics[height=3cm]{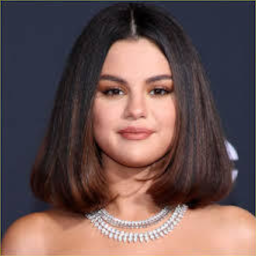}\end{minipage}
    & is a portrait photograph of a famous person. She is wearing two necklaces. She has dark hair and is wearing makeup. She is facing the camera and the background is black. \\
    \begin{minipage}{3.5cm}\includegraphics[height=3cm]{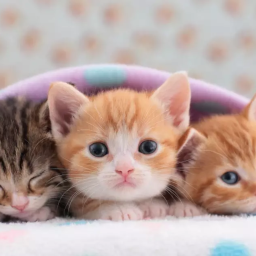}\end{minipage}
    & is a cute photograph of three kittens. They are under a blanket. The background is blurred but it seems white and orange. The blanket is purple. The two cats on the right are orange and the one on the left is grey. The orange cats have open eyes and the grey cat has closed eyes. They are all super cute. \\
    \begin{minipage}{3.5cm}
    \begin{tikzpicture}
        \node[anchor=center,inner sep=0] (image) at (0,0) {\includegraphics[height=3cm,width=3cm]{example-image-plain}};
        \node[black!80] at (0, 0) {\footnotesize Image to caption};
    \end{tikzpicture}
    \end{minipage}
    & is a realistic photograph of a \texttt{[label name]}. \texttt{[...]}
\end{tabular}

These images are not from \imagenet but downloaded from the web.
Choosing these captions is only done once and then fixed for all experiments in the paper.
We also set the decoding strategy to be \emph{greedy} (as we did not observe significant improvements from using beam search).
We highlight that any expert knowledge, if needed, is only required to annotate these three images.
Compared to annotating a full test set, the cost is negligible.

\clearpage
\section{Additional results}

\subsection{Open-ended failure search}\label{sec:additional_openended}

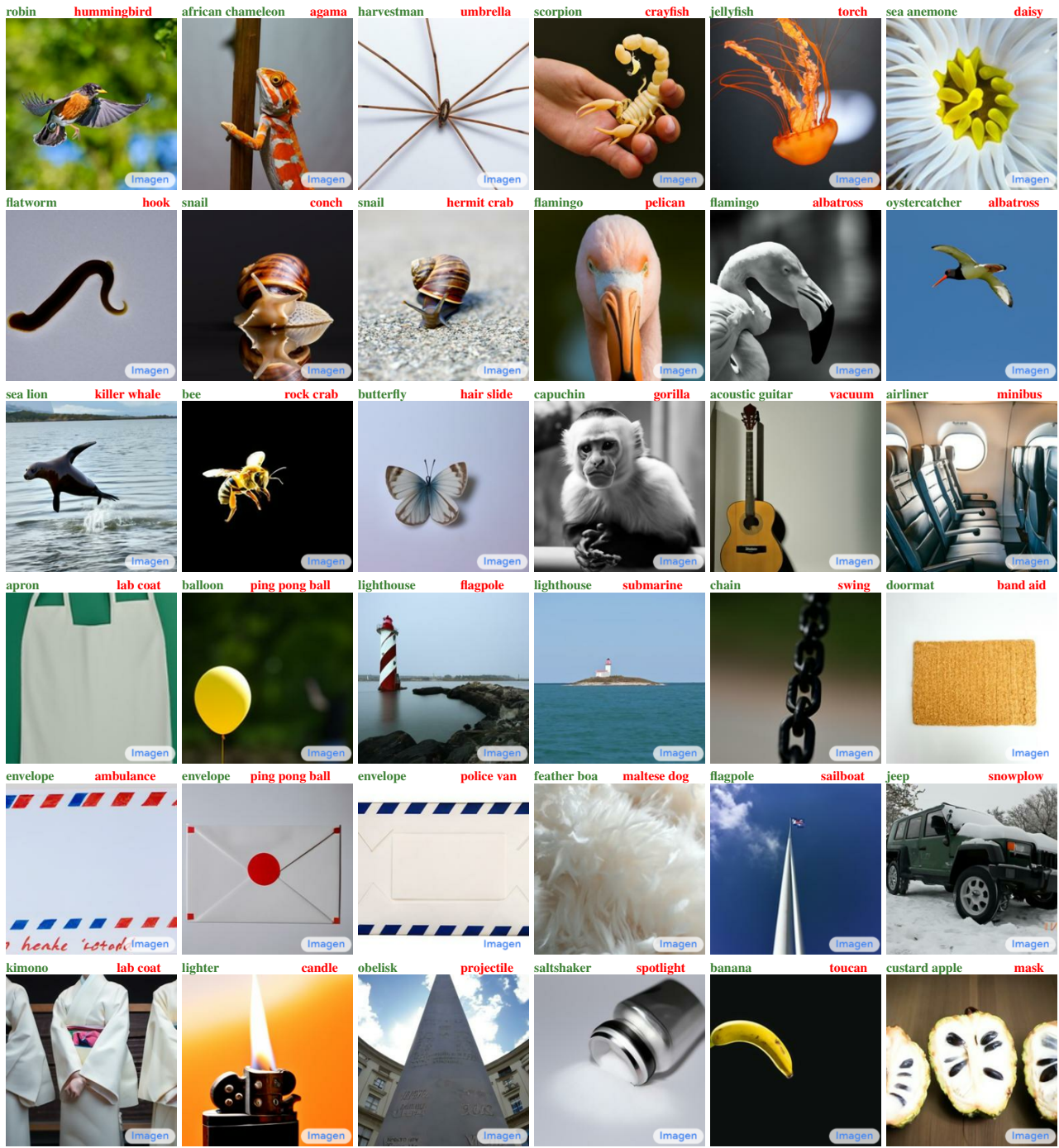
\begin{figure}[h]
\centering
\begin{overpic}[width=0.160\linewidth]{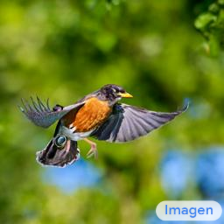}
\put(0,102){\tiny \bf {\color{OliveGreen} robin}}
\put(40,102){\tiny \bf {\color{red} hummingbird}}
\end{overpic}
\begin{overpic}[width=0.160\linewidth]{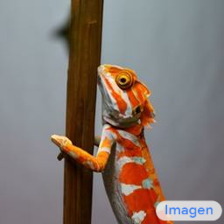}
\put(0,102){\tiny \bf {\color{OliveGreen} african chameleon}}
\put(75,102){\tiny \bf {\color{red} agama}}
\end{overpic}
\begin{overpic}[width=0.160\linewidth]{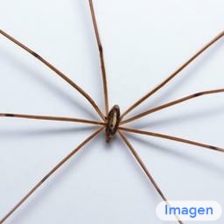}
\put(0,102){\tiny \bf {\color{OliveGreen} harvestman}}
\put(60,102){\tiny \bf {\color{red} umbrella}}
\end{overpic}
\begin{overpic}[width=0.160\linewidth]{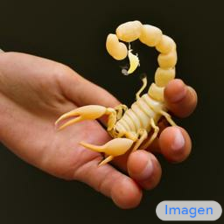}
\put(0,102){\tiny \bf {\color{OliveGreen} scorpion}}
\put(65,102){\tiny \bf {\color{red} crayfish}}
\end{overpic}
\begin{overpic}[width=0.160\linewidth]{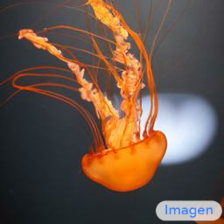}
\put(0,102){\tiny \bf {\color{OliveGreen} jellyfish}}
\put(75,102){\tiny \bf {\color{red} torch}}
\end{overpic}
\begin{overpic}[width=0.160\linewidth]{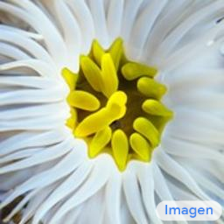}
\put(0,102){\tiny \bf {\color{OliveGreen} sea anemone}}
\put(75,102){\tiny \bf {\color{red} daisy}}
\end{overpic}
\\ \vspace{3mm}
\begin{overpic}[width=0.160\linewidth]{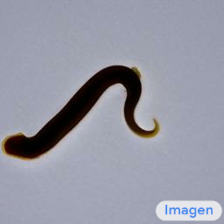}
\put(0,102){\tiny \bf {\color{OliveGreen} flatworm}}
\put(80,102){\tiny \bf {\color{red} hook}}
\end{overpic}
\begin{overpic}[width=0.160\linewidth]{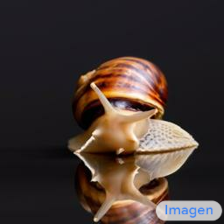}
\put(0,102){\tiny \bf {\color{OliveGreen} snail}}
\put(75,102){\tiny \bf {\color{red} conch}}
\end{overpic}
\begin{overpic}[width=0.160\linewidth]{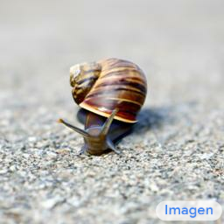}
\put(0,102){\tiny \bf {\color{OliveGreen} snail}}
\put(52,102){\tiny \bf {\color{red} hermit crab}}
\end{overpic}
\begin{overpic}[width=0.160\linewidth]{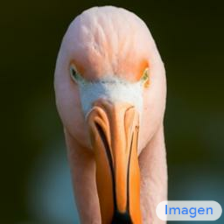}
\put(0,102){\tiny \bf {\color{OliveGreen} flamingo}}
\put(65,102){\tiny \bf {\color{red} pelican}}
\end{overpic}
\begin{overpic}[width=0.160\linewidth]{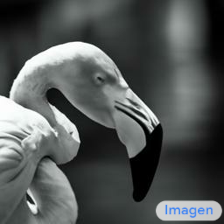}
\put(0,102){\tiny \bf {\color{OliveGreen} flamingo}}
\put(60,102){\tiny \bf {\color{red} albatross}}
\end{overpic}
\begin{overpic}[width=0.160\linewidth]{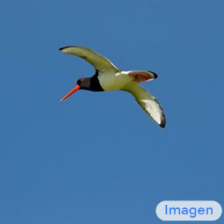}
\put(0,102){\tiny \bf {\color{OliveGreen} oystercatcher}}
\put(60,102){\tiny \bf {\color{red} albatross}}
\end{overpic}
\\ \vspace{3mm}
\begin{overpic}[width=0.160\linewidth]{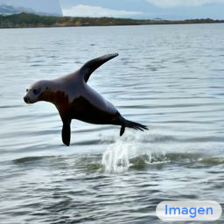}
\put(0,102){\tiny \bf {\color{OliveGreen} sea lion}}
\put(52,102){\tiny \bf {\color{red} killer whale}}
\end{overpic}
\begin{overpic}[width=0.160\linewidth]{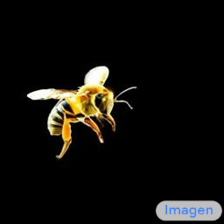}
\put(0,102){\tiny \bf {\color{OliveGreen} bee}}
\put(60,102){\tiny \bf {\color{red} rock crab}}
\end{overpic}
\begin{overpic}[width=0.160\linewidth]{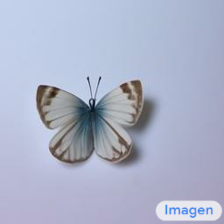}
\put(0,102){\tiny \bf {\color{OliveGreen} butterfly}}
\put(60,102){\tiny \bf {\color{red} hair slide}}
\end{overpic}
\begin{overpic}[width=0.160\linewidth]{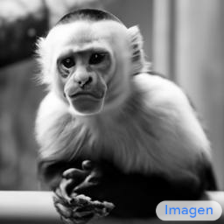}
\put(0,102){\tiny \bf {\color{OliveGreen} capuchin}}
\put(70,102){\tiny \bf {\color{red} gorilla}}
\end{overpic}
\begin{overpic}[width=0.160\linewidth]{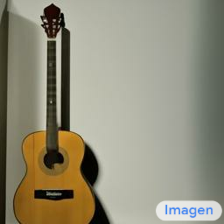}
\put(0,102){\tiny \bf {\color{OliveGreen} acoustic guitar}}
\put(70,102){\tiny \bf {\color{red} vacuum}}
\end{overpic}
\begin{overpic}[width=0.160\linewidth]{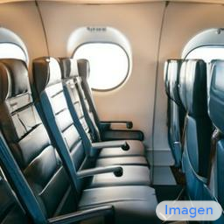}
\put(0,102){\tiny \bf {\color{OliveGreen} airliner}}
\put(65,102){\tiny \bf {\color{red} minibus}}
\end{overpic}
\\ \vspace{3mm}
\begin{overpic}[width=0.160\linewidth]{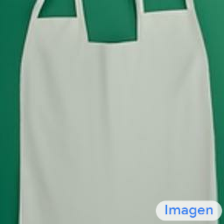}
\put(0,102){\tiny \bf {\color{OliveGreen} apron}}
\put(65,102){\tiny \bf {\color{red} lab coat}}
\end{overpic}
\begin{overpic}[width=0.160\linewidth]{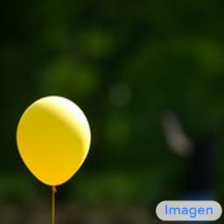}
\put(0,102){\tiny \bf {\color{OliveGreen} balloon}}
\put(40,102){\tiny \bf {\color{red} ping pong ball}}
\end{overpic}
\begin{overpic}[width=0.160\linewidth]{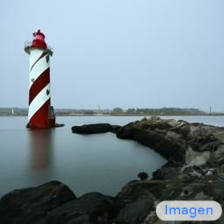}
\put(0,102){\tiny \bf {\color{OliveGreen} lighthouse}}
\put(60,102){\tiny \bf {\color{red} flagpole}}
\end{overpic}
\begin{overpic}[width=0.160\linewidth]{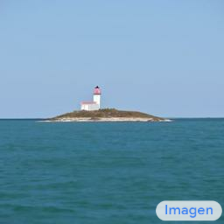}
\put(0,102){\tiny \bf {\color{OliveGreen} lighthouse}}
\put(52,102){\tiny \bf {\color{red} submarine}}
\end{overpic}
\begin{overpic}[width=0.160\linewidth]{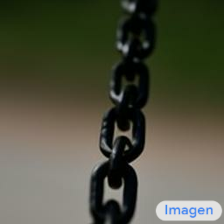}
\put(0,102){\tiny \bf {\color{OliveGreen} chain}}
\put(75,102){\tiny \bf {\color{red} swing}}
\end{overpic}
\begin{overpic}[width=0.160\linewidth]{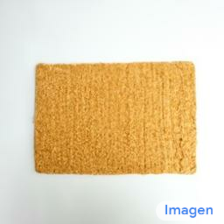}
\put(0,102){\tiny \bf {\color{OliveGreen} doormat}}
\put(65,102){\tiny \bf {\color{red} band aid}}
\end{overpic}
\\ \vspace{3mm}
\begin{overpic}[width=0.160\linewidth]{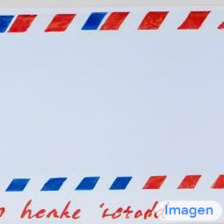}
\put(0,102){\tiny \bf {\color{OliveGreen} envelope}}
\put(52,102){\tiny \bf {\color{red} ambulance}}
\end{overpic}
\begin{overpic}[width=0.160\linewidth]{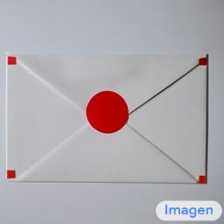}
\put(0,102){\tiny \bf {\color{OliveGreen} envelope}}
\put(40,102){\tiny \bf {\color{red} ping pong ball}}
\end{overpic}
\begin{overpic}[width=0.160\linewidth]{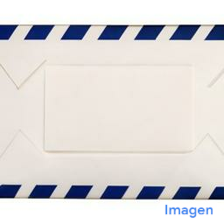}
\put(0,102){\tiny \bf {\color{OliveGreen} envelope}}
\put(60,102){\tiny \bf {\color{red} police van}}
\end{overpic}
\begin{overpic}[width=0.160\linewidth]{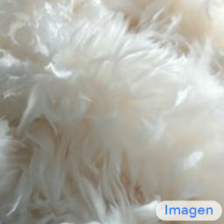}
\put(0,102){\tiny \bf {\color{OliveGreen} feather boa}}
\put(52,102){\tiny \bf {\color{red} maltese dog}}
\end{overpic}
\begin{overpic}[width=0.160\linewidth]{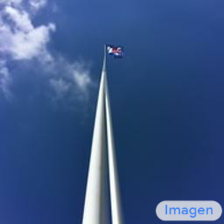}
\put(0,102){\tiny \bf {\color{OliveGreen} flagpole}}
\put(65,102){\tiny \bf {\color{red} sailboat}}
\end{overpic}
\begin{overpic}[width=0.160\linewidth]{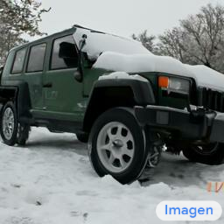}
\put(0,102){\tiny \bf {\color{OliveGreen} jeep}}
\put(60,102){\tiny \bf {\color{red} snowplow}}
\end{overpic}
\\ \vspace{3mm}
\begin{overpic}[width=0.160\linewidth]{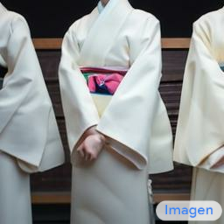}
\put(0,102){\tiny \bf {\color{OliveGreen} kimono}}
\put(65,102){\tiny \bf {\color{red} lab coat}}
\end{overpic}
\begin{overpic}[width=0.160\linewidth]{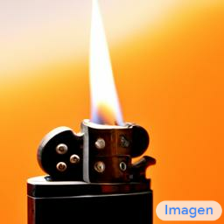}
\put(0,102){\tiny \bf {\color{OliveGreen} lighter}}
\put(70,102){\tiny \bf {\color{red} candle}}
\end{overpic}
\begin{overpic}[width=0.160\linewidth]{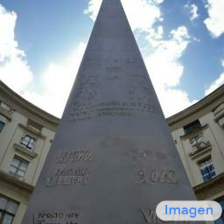}
\put(0,102){\tiny \bf {\color{OliveGreen} obelisk}}
\put(60,102){\tiny \bf {\color{red} projectile}}
\end{overpic}
\begin{overpic}[width=0.160\linewidth]{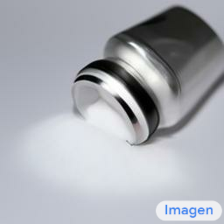}
\put(0,102){\tiny \bf {\color{OliveGreen} saltshaker}}
\put(60,102){\tiny \bf {\color{red} spotlight}}
\end{overpic}
\begin{overpic}[width=0.160\linewidth]{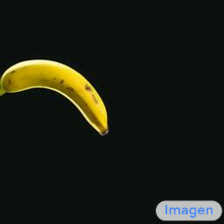}
\put(0,102){\tiny \bf {\color{OliveGreen} banana}}
\put(70,102){\tiny \bf {\color{red} toucan}}
\end{overpic}
\begin{overpic}[width=0.160\linewidth]{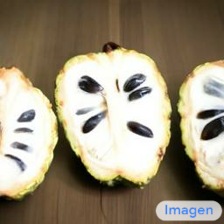}
\put(0,102){\tiny \bf {\color{OliveGreen} custard apple}}
\put(75,102){\tiny \bf {\color{red} mask}}
\end{overpic}
\caption{{\bf Illustration of failure cases listed in \autoref{tab:additional_openended}}. The correct label is to the left in {\color{OliveGreen} \bf green}. The incorrect prediction is to the right in {\color{red} \bf  red}. The model used is a \resnet-50 found on \tfhub. \label{fig:additional_openended}\vspace{-2cm}}
\end{figure}

\paragraph{Large scale experiments.} Similarly to \autoref{fig:openended_failures} and \autoref{table:openended_failures}, \autoref{fig:additional_openended} and \autoref{tab:additional_openended} show failure cases automatically found by our pipeline for a \resnet-50 available on \tfhub in a fully open-ended manner (i.e., without leveraging an external dataset).
The labels considered are a subset of the 200 labels present in \imageneta.
We let the reader interpret these failure cases themselves.
The failures are diverse and are due to different factors, such as:
\begin{enumerate*}[label={\it(\roman*)}]
    \item misleading color patterns (e.g., \texttt{sea amemone} $\rightarrow$ \texttt{daisy}),
    \item spurious context (e.g., \texttt{jeep} $\rightarrow$ \texttt{snowplow}),
    \item missing knowledge (e.g., \texttt{custard apple} $\rightarrow$ \texttt{mask}), or
    \item hallucinations (e.g., \texttt{feather boa} $\rightarrow$ \texttt{maltese dog}).
\end{enumerate*}

\begin{table}[t]
\begin{center}
\resizebox{.95\textwidth}{!}{
\begin{tabular}{ll|p{12cm}|lr}
True label & Target label & Caption & \multicolumn{2}{l}{Failure rate (target)} \\
\toprule
\multirow{2}{*}{\texttt{robin}} & \multirow{2}{*}{\texttt{hummingbird}} & a realistic photograph of a robin (oscine). & 0.0032\% & 1$\times$ \\
& & \dittoclosing It is flying. & 7.35\% & 2264.3$\times$ \\
\hline
\multirow{2}{*}{\texttt{african chameleon}} & \multirow{2}{*}{\texttt{agama}} & a realistic photograph of an african chameleon (lizard). & 0.15\% & 1$\times$ \\
& & \dittoclosing He is holding a stick. The chameleon is orange and white. & 1.01\% & 6.7$\times$ \\
\hline
\multirow{2}{*}{\texttt{harvestman}} & \multirow{2}{*}{\texttt{umbrella}} & a realistic photograph of a harvestman (arthropod). & 0.45\% & 1$\times$ \\
& & \dittoclosing It is shot from above. The harvestman is on a white background. & 1.32\% & 2.9$\times$ \\
\hline
\multirow{2}{*}{\texttt{scorpion}} & \multirow{2}{*}{\texttt{crayfish}} & a realistic photograph of a scorpion (arthropod). & 0.0042\% & 1$\times$ \\
& & \dittoclosing It is on a person's hand. & 0.13\% & 29.5$\times$ \\
\hline
\multirow{2}{*}{\texttt{jellyfish}} & \multirow{2}{*}{\texttt{torch}} & a realistic photograph of a jellyfish (invertebrate). & 0.14\% & 1$\times$ \\
& & \dittoclosing The background is black. The jellyfish is orange. & 0.45\% & 3.2$\times$ \\
\hline
\multirow{2}{*}{\texttt{sea anemone}} & \multirow{2}{*}{\texttt{daisy}} & a realistic photograph of a sea anemone (coelenterate). & 0.32\% & 1$\times$ \\
& & \dittoclosing It is yellow and white. The background is blurred. & 1.33\% & 4.2$\times$ \\
\hline
\multirow{2}{*}{\texttt{flatworm}} & \multirow{2}{*}{\texttt{hook}} & a realistic photograph of a flatworm (invertebrate). & 0.61\% & 1$\times$ \\
& & \dittoclosing It is on a white background. & 1.58\% & 2.6$\times$ \\
\hline
\multirow{2}{*}{\texttt{snail}} & \multirow{2}{*}{\texttt{conch}} & a realistic photograph of a snail (mollusk). & 0.039\% & 1$\times$ \\
& & \dittoclosing It is on a black background. The snail is reflected on the floor. & 0.88\% & 22.5$\times$ \\
\hline
\multirow{2}{*}{\texttt{snail}} & \multirow{2}{*}{\texttt{hermit crab}} & a realistic photograph of a snail (mollusk). & 0.0082\% & 1$\times$ \\
& & \dittoclosing It is on a grey road. & 0.10\% & 12.2$\times$ \\
\hline
\multirow{2}{*}{\texttt{flamingo}} & \multirow{2}{*}{\texttt{pelican}} & a realistic photograph of a flamingo (aquatic bird). & 0.023\% & 1$\times$ \\
& & \dittoclosing It is a close up of the head. The flamingo is facing the camera. & 0.87\% & 37.6$\times$ \\
\hline
\multirow{2}{*}{\texttt{flamingo}} & \multirow{2}{*}{\texttt{albatross}} & a realistic photograph of a flamingo (aquatic bird). & 0.081\% & 1$\times$ \\
& & \dittoclosing It is black and white. The flamingo is looking to the right. & 1.30\% & 16.1$\times$ \\
\hline
\multirow{2}{*}{\texttt{oystercatcher}} & \multirow{2}{*}{\texttt{albatross}} & a realistic photograph of an oystercatcher (wading bird). & 0.0025\% & 1$\times$ \\
& & \dittoclosing It is flying. & 0.52\% & 208.2$\times$ \\
\hline
\multirow{2}{*}{\texttt{sea lion}} & \multirow{2}{*}{\texttt{killer whale}} & a realistic photograph of a sea lion (seal). & 0.14\% & 1$\times$ \\
& & \dittoclosing It is jumping out of the water. & 4.31\% & 31.3$\times$ \\
\hline
\multirow{2}{*}{\texttt{bee}} & \multirow{2}{*}{\texttt{rock crab}} & a realistic photograph of a bee (insect). & 0.086\% & 1$\times$ \\
& & \dittoclosing It is flying. The background is black. & 0.18\% & 2.1$\times$ \\
\hline
\multirow{2}{*}{\texttt{cabbage butterfly}} & \multirow{2}{*}{\texttt{hair slide}} & a realistic photograph of a cabbage butterfly (butterfly). & 0.099\% & 1$\times$ \\
& & \dittoclosing It is on a white background. It is in the middle of the image. & 1.98\% & 20.0$\times$ \\
\hline
\multirow{2}{*}{\texttt{capuchin}} & \multirow{2}{*}{\texttt{gorilla}} & a realistic photograph of a capuchin (monkey). & 0.011\% & 1$\times$ \\
& & \dittoclosing It is a black and white photograph. & 0.84\% & 73.9$\times$ \\
\hline
\multirow{2}{*}{\texttt{acoustic guitar}} & \multirow{2}{*}{\texttt{vacuum}} & a realistic photograph of an acoustic guitar (stringed instrument). & 0.20\% & 1$\times$ \\
& & \dittoclosing It is leaning against a wall. & 1.02\% & 5.0$\times$ \\
\hline
\multirow{2}{*}{\texttt{airliner}} & \multirow{2}{*}{\texttt{minibus}} & a realistic photograph of an airliner (heavier-than-air craft). & 0.16\% & 1$\times$ \\
& & \dittoclosing There are seats in the foreground. & 3.05\% & 19.2$\times$ \\
\hline
\multirow{2}{*}{\texttt{apron}} & \multirow{2}{*}{\texttt{lab coat}} & a realistic photograph of an apron (clothing). & 0.44\% & 1$\times$ \\
& & \dittoclosing It is white. & 3.57\% & 8.1$\times$ \\
\hline
\multirow{2}{*}{\texttt{balloon}} & \multirow{2}{*}{\texttt{ping-pong ball}} & a realistic photograph of a balloon (aircraft). & 0.53\% & 1$\times$ \\
& & \dittoclosing It is yellow. The background is blurred. & 17.86\% & 33.8$\times$ \\
\hline
\multirow{2}{*}{\texttt{lighthouse}} & \multirow{2}{*}{\texttt{flagpole}} & a realistic photograph of a beacon (structure). & 0.095\% & 1$\times$ \\
& & \dittoclosing The lighthouse has red and white stripes. & 2.12\% & 22.4$\times$ \\
\hline
\multirow{2}{*}{\texttt{lighthouse}} & \multirow{2}{*}{\texttt{submarine}} & a realistic photograph of a beacon (structure). & 0.041\% & 1$\times$ \\
& & \dittoclosing It is on a small island at the horizon. & 12.5\% & 308.0$\times$ \\
\hline
\multirow{2}{*}{\texttt{chain}} & \multirow{2}{*}{\texttt{swing}} & a realistic photograph of a chain (attachment). & 1.22\% & 1$\times$ \\
& & \dittoclosing The chain is vertical. The chain is in focus. & 12.5\% & 10.2$\times$ \\
\hline
\multirow{2}{*}{\texttt{doormat}} & \multirow{2}{*}{\texttt{band aid}} & a realistic photograph of a doormat (floor cover). & 0.94\% & 1$\times$ \\
& & \dittoclosing The doormat is rectangular and is on a white background. & 5.68\% & 6.1$\times$ \\
\hline
\multirow{2}{*}{\texttt{envelope}} & \multirow{2}{*}{\texttt{ambulance}} & a realistic photograph of an envelope (instrumentality). & 0.210\% & 1$\times$ \\
& & \dittoclosing It has white and has red and blue stripes at the top and bottom. & 17.86\% & 60.0$\times$ \\
\hline
\multirow{2}{*}{\texttt{envelope}} & \multirow{2}{*}{\texttt{ping-pong ball}} & a realistic photograph of an envelope (instrumentality). & 1.04\% & 1$\times$ \\
& & \dittoclosing It is white and has a red dot on it. & 75.\% & 72.0$\times$ \\
\hline
\multirow{2}{*}{\texttt{envelope}} & \multirow{2}{*}{\texttt{police van}} & a realistic photograph of an envelope (instrumentality). & 0.510\% & 1$\times$ \\
& & \dittoclosing It has white and has white and blue diagonal stripes at the top and bottom. & 6.94\% & 11.7$\times$ \\
\hline
\multirow{2}{*}{\texttt{feather boa}} & \multirow{2}{*}{\texttt{maltese dog}} & a realistic photograph of a feather boa (garment). & 4.59\% & 1$\times$ \\
& & \dittoclosing It is white and fluffy. & 41.67\% & 9.1$\times$ \\
\hline
\multirow{2}{*}{\texttt{flagpole}} & \multirow{2}{*}{\texttt{sailboat}} & a realistic photograph of a flagpole (stick). & 0.19\% & 1$\times$ \\
& & \dittoclosing It is white and the sky is blue. & 2.19\% & 11.5$\times$ \\
\hline
\multirow{2}{*}{\texttt{jeep}} & \multirow{2}{*}{\texttt{snowplow}} & a realistic photograph of a jeep (motor vehicle). & 0.30\% & 1$\times$ \\
& & \dittoclosing It is parked in the snow. & 17.86\% & 59.3$\times$ \\
\hline
\multirow{2}{*}{\texttt{kimono}} & \multirow{2}{*}{\texttt{lab coat}} & a realistic photograph of a kimono (garment). & 0.69\% & 1$\times$ \\
& & \dittoclosing It is white. & 2.84\% & 4.1$\times$ \\
\hline
\multirow{2}{*}{\texttt{lighter}} & \multirow{2}{*}{\texttt{candle}} & a realistic photograph of a lighter (instrumentality). & 5.37\% & 1$\times$ \\
& & \dittoclosing It has a flame coming out of it. & 41.67\% & 7.8$\times$ \\
\hline
\multirow{2}{*}{\texttt{obelisk}} & \multirow{2}{*}{\texttt{projectile}} & a realistic photograph of an obelisk (structure). & 0.14\% & 1$\times$ \\
& & \dittoclosing It is pointing up. The sky is blue. & 1.09\% & 7.7$\times$ \\
\hline
\multirow{3}{*}{\texttt{saltshaker}} & \multirow{3}{*}{\texttt{spotlight}} & a realistic photograph of a saltshaker (container). & 1.02\% & 1$\times$ \\
& & \dittoclosing It has a silver lid. The salt shaker is on a white background. The salt is spilling out of the jar. & 13.89\% & 13.7$\times$ \\
\hline
\multirow{2}{*}{\texttt{banana}} & \multirow{2}{*}{\texttt{toucan}} & a realistic photograph of a banana (produce). & 0.0058\% & 1$\times$ \\
& & \dittoclosing It is yellow and is floating in the air. The background is black. & 0.047\% & 8.2$\times$ \\
\hline
\multirow{2}{*}{\texttt{custard apple}} & \multirow{2}{*}{\texttt{mask}} & a realistic photograph of a custard apple (produce). & 0.32\% & 1$\times$ \\
& & \dittoclosing The fruit is cut in half. & 4.46\% & 14.0$\times$ \\
\bottomrule
\end{tabular}
}
\end{center}
\caption{\textbf{Absolute failure rates of a \resnet-50 for 36 additional true and target label pairs.} We show the target failure rate (i.e., the model prediction is the target label). Captions are automatically discovered using the method detailed in \autoref{sec:method}. Note that to the contrary of \autoref{table:openended_failures}, we consider an image to be misclassified when the top-1 prediction is wrong (and not from the same \wordnet parent) rather than when the true label is not part of the top-3 predictions. \label{tab:additional_openended}}
\end{table}

\clearpage

\paragraph{Generalization to other vision architectures.}
We investigate whether our approach extends to other, more challenging backbones, such as a \vit{-B/32} and \vit{-B/8} which obtain much better performance than a \resnet{50} on \imagenet.
We use the \texttt{fly} and \texttt{bee} failure case as our case study and run the same experiment as the one presented in \autoref{table:openended_failures}.
Results are reported in \autoref{tab:generalizationtovits}.
First, we observe that both \vit models exhibit the same bias than the one found for the \resnet{-50} model (that a fly on a flower is more often confused as a bee).
Second, while the failure rates increase significantly compared to the baseline, the bias seems to be less pronounced (with only a 114$\times$ and 13.5$\times$ increase in failure rates compared to 497$\times$ for the \resnet) which highlights the qualities of both models.

\begin{table}[h]
\begin{center}
\resizebox{\textwidth}{!}{
\begin{tabular}{lll|p{8cm}|lr}
Architecture & True label & Target label & Caption & \multicolumn{2}{l}{Failure rate (target)} \\
\toprule
\multirow{2}{*}{\vit{-B/32}} & \multirow{2}{*}{\texttt{fly}} & \multirow{2}{*}{\texttt{bee}} & a realistic photograph of a fly (insect). & 0.0002\% & 1$\times$ \\
& & & \dittoclosing it is on a flower. & 0.02278\% & 113.9$\times$ \\
\hline
\multirow{2}{*}{\vit{-B/8}} & \multirow{2}{*}{\texttt{fly}} & \multirow{2}{*}{\texttt{bee}} & a realistic photograph of a fly (insect). & 0.0002\% & 1$\times$ \\
& & & \dittoclosing it is on a flower. & 0.0027\% & 13.5$\times$ \\
\bottomrule
\end{tabular}
}
\end{center}
\caption{\textbf{Absolute failure rates of a \vit-B/8, \vit-B/32 on the \texttt{fly}/\texttt{bee} failure case.} We show the target failure rate (i.e., the model prediction is the target label). Captions are automatically discovered using the method detailed in \autoref{sec:method}.  \label{tab:generalizationtovits}}
\end{table}

\paragraph{Generalization to images from Google Image Search.}
We expand on the analysis in the main text on how failures transfer to images downloaded through {\em Google Image Search} by exploring results on additional prompts.
For each prompt in \autoref{tab:additional_openended}, we query {\em Google Image Search} and automatically download the first 100 images (if we download too many images, the later, less relevant images may fail to capture the prompt).
We take five pretrained \resnet-50s and measure the confusion rate for the target class on these images in \autoref{tab:generalisetorealimages}. 
The results demonstrate that the failures we found with generated data generalize to real images.
For the other prompts (not shown in \autoref{tab:generalisetorealimages}), we found that the target failure rate is zero both for the original and modified prompt.
This is presumably because there were not enough images to surface the failure rate. %
This demonstrates an additional benefit of our approach: we are not limited by images that exist on the web or are retrievable through an image search engine. 

\begin{table}[h]
\begin{center}
\resizebox{\textwidth}{!}{
\begin{tabular}{ll|p{9cm}|l}
True label & Target label & Caption  & {Failure rate (target)}\\
\toprule
\multirow{2}{*}{\texttt{robin}} & \multirow{2}{*}{\texttt{hummingbird}} & a realistic photograph of a robin (oscine).  & 1.10\% $\pm$ 1.10\\
& & \dittoclosing It is flying.  & 10.31\% $\pm$ 3.85\\
\hline
\multirow{2}{*}{\texttt{african chameleon}} & \multirow{2}{*}{\texttt{agama}} & a realistic photograph of an african chameleon (lizard).  & 1.70\% $\pm$ 0.59\\
& & \dittoclosing He is holding a stick. The chameleon is orange and white.  & 4.35\% $\pm$ 0.77\\
\hline
\multirow{2}{*}{\texttt{scorpion}} & \multirow{2}{*}{\texttt{crayfish}} & a realistic photograph of a scorpion (arthropod).  & 0.21\% $\pm$ 0.46\\
& & \dittoclosing It is on a person's hand.  & 0.67\% $\pm$ 1.01\\
\hline
\multirow{2}{*}{\texttt{lighthouse}} & \multirow{2}{*}{\texttt{submarine}} & a realistic photograph of a beacon (structure).  & 0.38\% $\pm$ 0.85\\
& & \dittoclosing It is on a small island at the horizon. & 2.68\% $\pm$ 0.56\\
\hline
\multirow{2}{*}{\texttt{jeep}} & \multirow{2}{*}{\texttt{snowplow}} & a realistic photograph of a jeep (motor vehicle). & 5.84\% $\pm$ 1.23\\
& & \dittoclosing It is parked in the snow. & 21.89\% $\pm$ 4.30\\
\hline
\multirow{2}{*}{\texttt{lighter}} & \multirow{2}{*}{\texttt{candle}} & a realistic photograph of a lighter (instrumentality). & 0.65\% $\pm$ 0.96\\
& & \dittoclosing It has a flame coming out of it.  & 2.63\% $\pm$ 1.86\\
\hline
\end{tabular}
}
\end{center}
\caption{{\bf Failure rates for images from {\em Google Image Search}.} We report the mean and standard deviation of the target failure rate for the original and discovered captions over five randomly initialized \resnet-50s. We find that the failure rates go up with the modified caption if the failure rate is non zero initially. \label{tab:generalisetorealimages}}
\end{table}

\paragraph{Statistical significance of results.}
We further investigate the significance of the results on two of the open-ended failure cases in the main paper (the ones exhibiting larger failure rates).
Here, we evaluate our \resnet-50 and generate samples until we either find 10 images that cause the classifier to mispredict the class towards the target class (using top-1 accuracy instead of the typical top-3 to allow us to run many experiments efficiently) or find no misclassification towards the target class within 20K samples. We report the failure rate for the original and discovered captions and compute p-values using the Mann-Whitney U test \citep{mann1947test} at a significance level of 0.005 to determine if the differences in failure rates are statistically significant.
Results are in \autoref{tab:pvaluessignificance}.
As the p-values ($0.00015$, $6.34\cdot10^{-5}$)  are lower than $0.005$, we find a {\em significant} result that images of the discovered caption (e.g., ``... crayfish (crustacean). it is in a net.'') are more often misclassified for the target class (e.g., \texttt{chainlink fence}) than the original caption (e.g., ``... crayfish (crustacean).'').

\begin{table}[h]
\begin{center}
\resizebox{\textwidth}{!}{
\begin{tabular}{ll|p{7cm}|lr}
True label & Target label & Caption & {Failure rate (target)} & p-value \\
\toprule
\multirow{2}{*}{\texttt{fly}} & \multirow{2}{*}{\texttt{bee}} & a realistic photograph of a fly (insect). & $0.00\% \pm 0.00$  & \multirow{2}{*}{$0.00015$} \\
& & \dittoclosing it is on a flower. & $0.58\% \pm 0.15$ &  \\
\hline

\multirow{2}{*}{\texttt{crayfish}} & \multirow{2}{*}{\texttt{chainlink fence}} & a realistic photograph of a crayfish (crustacean). & $0.00\% \pm 0.00$ & \multirow{2}{*}{$6.34 \cdot 10^{-5}$} \\
& & \dittoclosing it is in a net. & $2.09\% \pm 0.56$ & \\
\bottomrule
\end{tabular}
}
\end{center}
\caption{{\bf Significance of failure rates.} We report the mean and standard deviation of the target failure rate for the original and discovered captions over ten runs. We then compute the p-value to determine if the difference in failure rates between the original and discovered caption is statistically significant. We do this for two failure cases and find that our results are statistically significant.  \label{tab:pvaluessignificance}}
\end{table}

\subsection{Additional comparisons on \ingresnet, \ingvit}
\label{app:transferstudy}

\paragraph{Models considered.} 
We collate a large set of models trained on \imagenet with differing size, pretraining, augmentation, and architectures in addition to the two \resnet{s} and \vit{s} we trained:
\squishlist
    \item \vit{-B*}, \vit{-L*}, \vit{-S*} \citep{dosovitskiy_image_2020}: \vit{s} pretrained on \imagenet{21K}.
    \item \vit{-R*} \citep{steiner2022train}: a hybrid \vit and \resnet model pretrained on \imagenet{21K}.
    \item  \bit{-*} \citep{kolesnikov2020big}: \bit models pretrained either on \imagenet{21K} (\bit{-M} *) or not pretrained (\bit{-S} *).
    \item \inception{\_}\resnet{ V2} \citep{szegedy2017inception}: a hybrid \inception, \resnet model with no pretraining.
    \item \inception{*} \citep{szegedy2015going}: \inception models with no pretraining.
    \item \resnet{*} \citep{he_deep_2015}: \resnet models with no pretraining.
\squishend

\begin{figure}[b]
  \caption{{\bf Failure rates (top-3) for different models on two generated datasets and \imagenet.} We report the failure rates of different models trained on \imagenet on both \ingresnet and \ingvit as well as \imagenet.}
  \label{fig:app:all_results_transfer}
  \includegraphics[width=\textwidth,trim={0cm 3cm 0cm 10cm},clip]{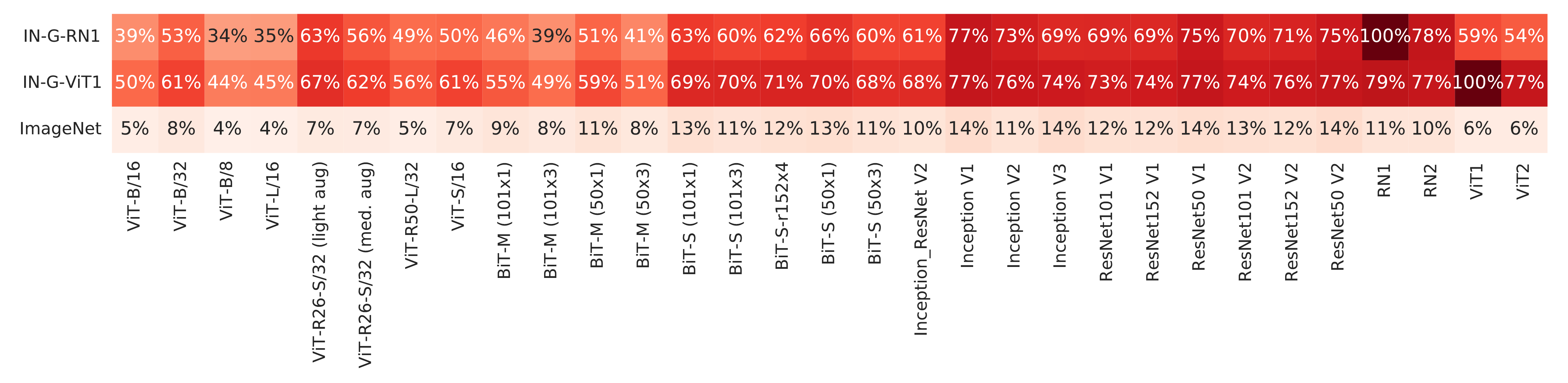}
\end{figure}

\paragraph{Transferability of errors on \ingresnet and \ingvit.}
We evaluate how often failures in \ingresnet and \ingvit transfer to these models in \autoref{fig:app:all_results_transfer}.
We can see that failures from both \ingresnet and \ingvit transfer across model architectures.
However, the \vit{s} and \bit{s} which are pretrained on \imagenet{21K} and achieve lower error on \imagenet are fooled the least often.
Within a model class, larger versions of the model seem more robust.
For example, \vit{-B/16} is more robust than \vit{-B/32} and similarly the larger \bit{s} (those of size 101x1) are more robust than their smaller counterparts (those of size 50x1).
Thus, stronger pretraining and larger models seem to lead to improved (but not complete) robustness against these generated datasets.

\begin{figure}[t]
  \caption{{\bf Error consistency for all models on the combined \ingresnet and \ingvit dataset.}}
  \label{fig:app:consistency}
  \includegraphics[width=\textwidth]{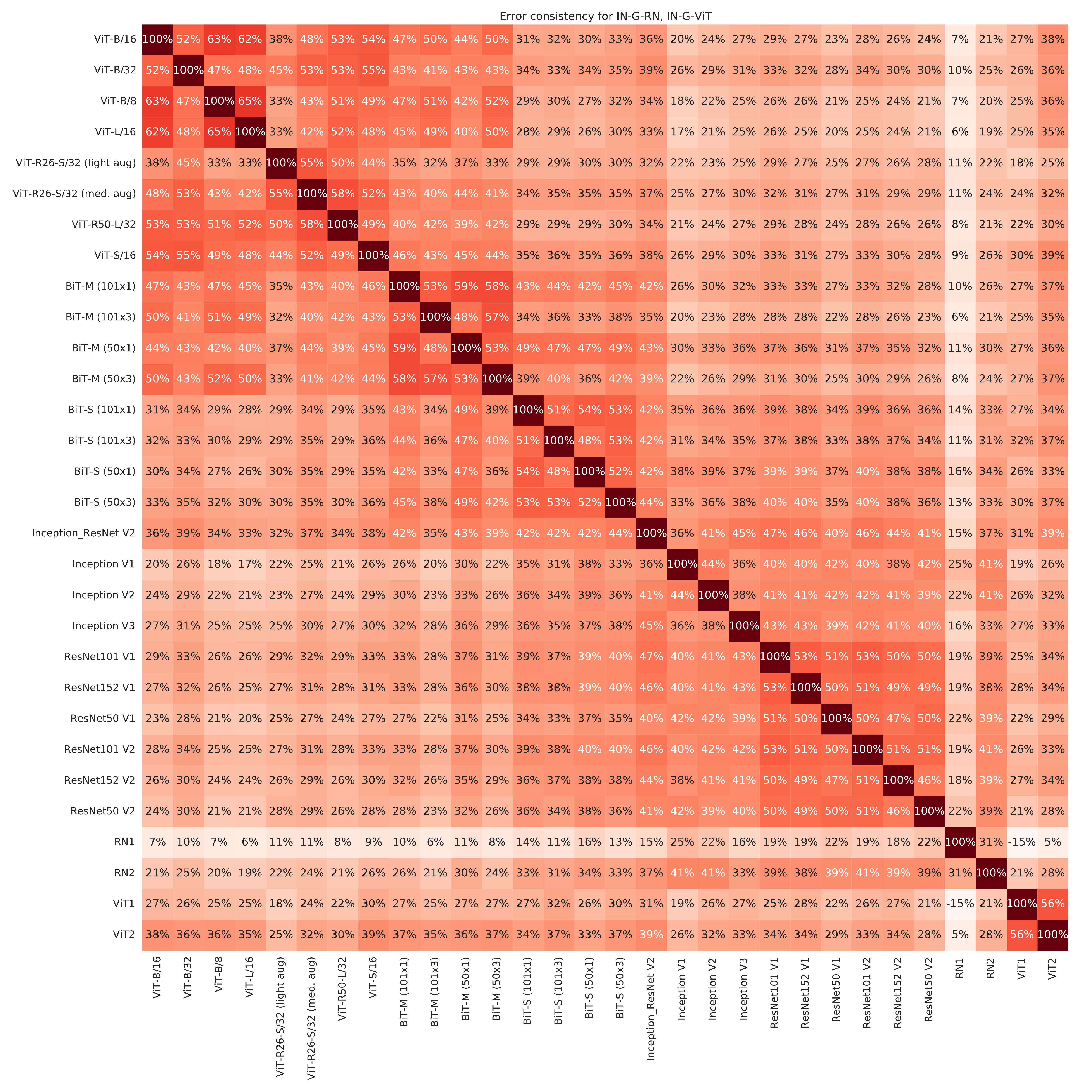}
\end{figure}

\paragraph{Error consistency on \ingresnet and \ingvit.}
Finally, we measure the error consistency of the models in \autoref{fig:app:consistency}.
We combine \ingresnet and \ingvit into one dataset and evaluate how often models make similar errors while accounting for the accuracy of each model (see Eq.~3 in \citealp{geirhos2021partial}).
A value of 100\% indicates that the errors two models make are perfectly correlated and -100\% that they are perfectly anti-correlated.
It is striking that errors are most consistent within a model class: \resnet{s} make similar errors to other \resnet{s} trained in a similar manner and similarly \bit{s} make similar errors to other \bit{s}, especially \bit{s} trained in the same manner.

\clearpage

\subsection{Reliability of the text-to-image model}
\label{app:imagenetaccuracy}
We measured the error rate of our text-to-image model on all 200 labels present in ImageNet-A by generating 10 images per class (totalling 2000 images). We note that in our pipeline we only consider egregious errors as wrong (where the classifier top-3 does not include the correct label or any label under the same WordNet parent). Of the generated images, 3.95\% did not represent the correct label and 1.45\% showed an item from the wrong WordNet family (e.g., asking for an ocarina sometimes generated a maraca - both are musical instruments but only the ocarina is a wind instrument). Only a single class (porcupine) was systematically misrepresented.
This experiment demonstrates that in general the text-to-image model is reliable and generates images of the right class.

\section{Interpreting failure cases}\label{sec:interpreting}

In this section, we aim at further characterizing failure cases by investigating why the models considered in our work yield wrong predictions on \ingresnet and \ingvit instances.
For that, we compared a few failure cases with their respective nearest neighbors within the training set of \imagenet in order to find patterns that shed light on the reasons behind wrong predictions. 

We find the ten nearest neighbors of an \ingvit instance in the embedding space induced by the second-to-last layer of a \vit trained on \imagenet, using cosine similarity as the distance measure.
This particular model achieves 82.7\% top-1 accuracy on the \imagenet validation set and has a failure rate of 100\% on \ingvit.

\begin{figure}[h]
\centering
\subfigure[\ingvit failure case.  \label{fig:in_g_mush_snail}]{\includegraphics[width=0.25\textwidth]{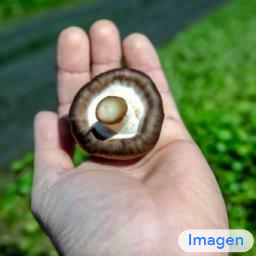}}
\subfigure[10 nearest neighbors of  \subref{fig:in_g_mush_snail} in the \imagenet train set. \label{fig:nn_mush_snail_all}]{\includegraphics[width=0.85\textwidth]{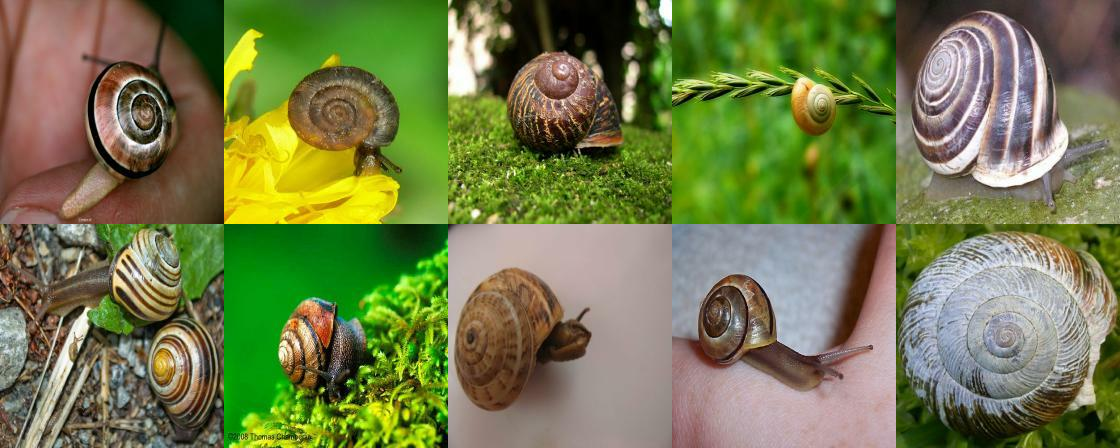}}
\subfigure[10 nearest neighbors of \subref{fig:in_g_mush_snail} in the \imagenet train set and in its respective label (\texttt{mushroom}). \label{fig:nn_mush_snail_same}]{\includegraphics[width=0.85\textwidth]{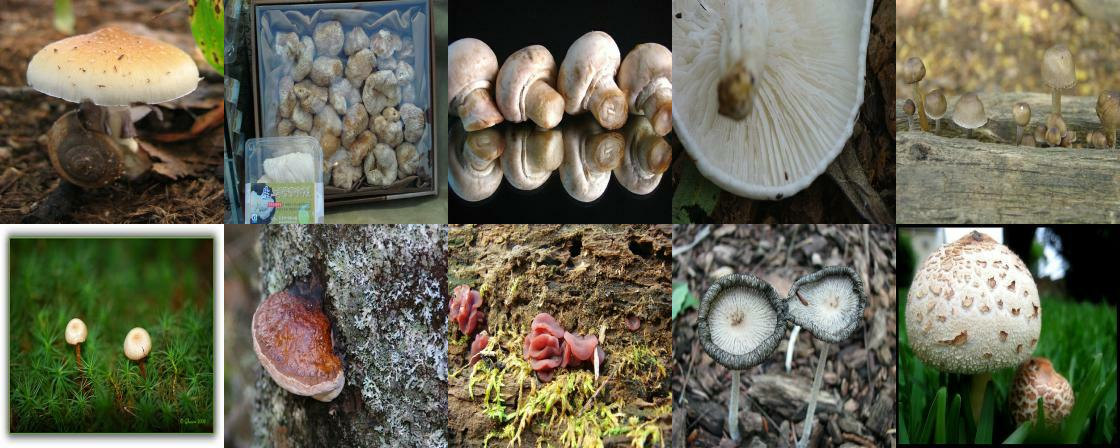}}
\caption{\textbf{Interpreting failure cases by inspecting nearest neighbors in the train set of \imagenet.} We analyze the failure case in \ingvit shown in panel \subref{fig:in_g_mush_snail}. The example is labeled as \texttt{mushroom} and classified as \texttt{snail} by a \vit trained on \imagenet. In panel~\subref{fig:nn_mush_snail_all}, we show the 10 nearest neighbors of \subref{fig:in_g_mush_snail} in the train set of \imagenet. All 10 neighbors are from the class \texttt{snail} and have similar features to the failure case, such as the background (e.g., the human hand), while the 10 nearest neighbors with the label \texttt{mushroom} showed in panel~\subref{fig:nn_mush_snail_same} do not have those features. This suggests that the \vit correlates such features with the label \texttt{snail}, and these  spurious correlations likely induced it to misclassify the image in \subref{fig:in_g_mush_snail}.}
\label{fig:nn_mush_as_snail}
\end{figure}

\begin{figure}[h]
\centering
\subfigure[\ingvit failure case.  \label{fig:in_g_butterfly_spider}]{\includegraphics[width=0.25\textwidth]{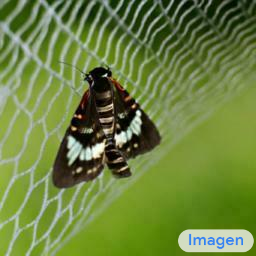}}
\subfigure[10 nearest neighbors of \subref{fig:in_g_butterfly_spider} in the \imagenet train set.\label{fig:nn_butterfly_spider_all}]{\includegraphics[width=0.85\textwidth]{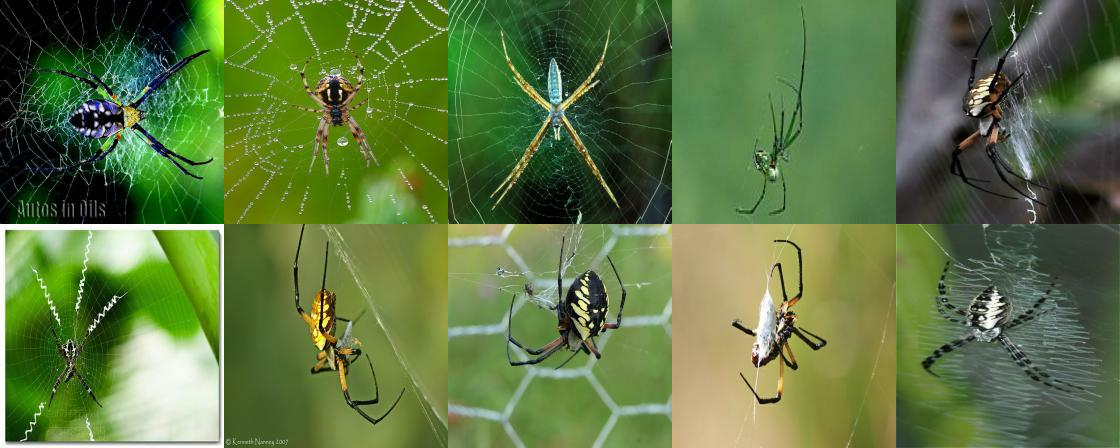}}
\subfigure[10 nearest neighbors of \subref{fig:in_g_butterfly_spider} in the \imagenet train set and in its respective class (\texttt{cabbage butterfly}). \label{fig:nn_butterfly_spider_same}]{\includegraphics[width=0.85\textwidth]{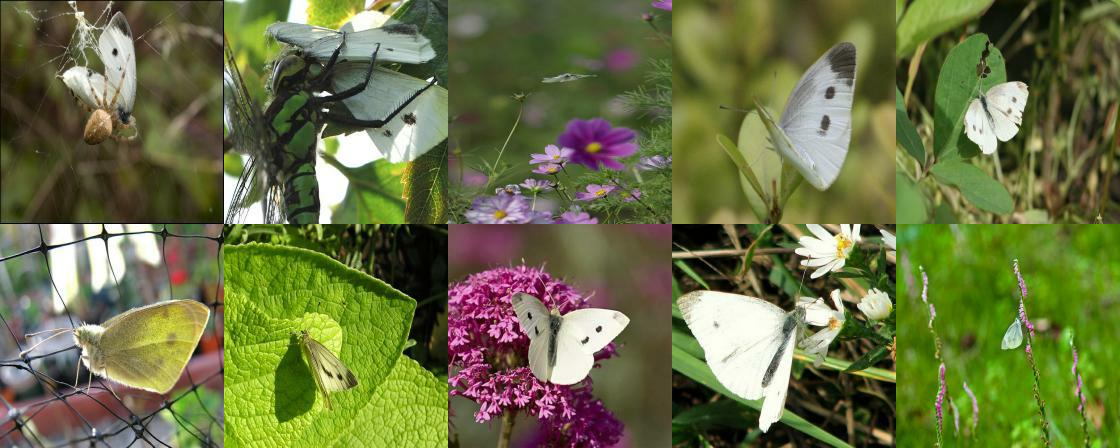}}
\caption{\textbf{Interpreting failure cases by inspecting their nearest neighbors in the train set of \imagenet.} 
We analyze the failure case in \ingvit shown in panel~\subref{fig:in_g_butterfly_spider}. The example is labeled as \texttt{cabbage butterfly} and classified as \texttt{barn spider} by a \vit trained on \imagenet. In panel~\subref{fig:nn_butterfly_spider_all} we show the 10 nearest neighbors of \subref{fig:in_g_butterfly_spider} in the train set of \imagenet. All 10 neighbors are from the classes \texttt{barn spider} or \texttt{black and gold garden spider} and have similarities to the failure case such as the white net in the background, while the 10 nearest neighbors in the class \texttt{cabbage butterfly} shown in panel~\subref{fig:nn_butterfly_spider_same} do not present those common features. This suggests that the \vit correlates such features with instances from the labels \texttt{barn spider} and \texttt{black and gold garden spider}, and exploiting these spurious correlations likely induced the model to misclassify the image in \subref{fig:in_g_butterfly_spider}.}
\label{fig:nn_butterfly_as_spider}
\end{figure}

\begin{figure}[h]
\centering
\subfigure[\ingvit failure case.  \label{fig:in_g_flag_ski}]{\includegraphics[width=0.25\textwidth]{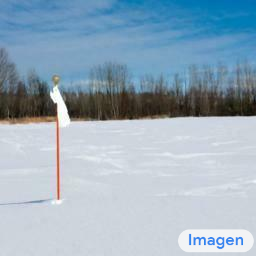}}
\subfigure[10 nearest neighbors of \subref{fig:in_g_flag_ski} in the \imagenet train set..  \label{fig:nn_flag_ski_all}]{\includegraphics[width=0.85\textwidth]{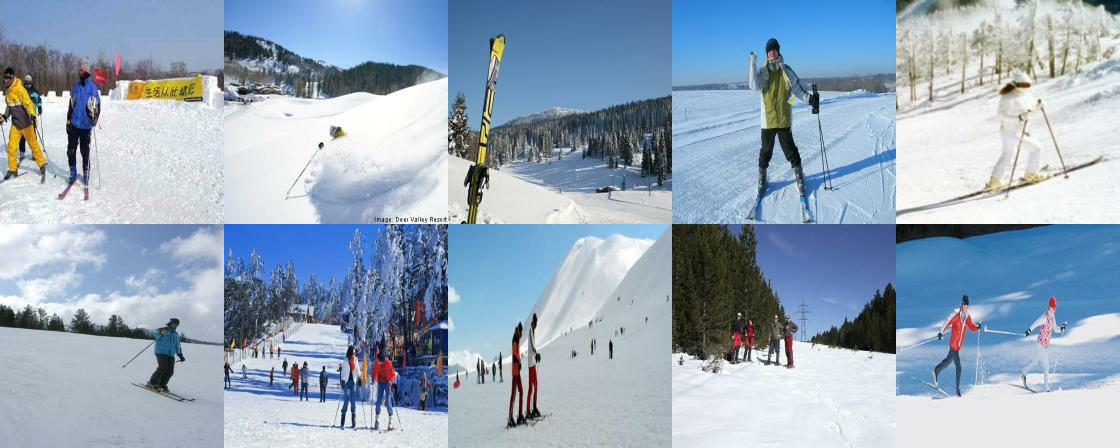}}
\subfigure[10 nearest neighbors of \subref{fig:in_g_flag_ski} in the \imagenet train set and in its respective class (\texttt{flagpole}) \label{fig:nn_flag_ski_same}]{\includegraphics[width=0.85\textwidth]{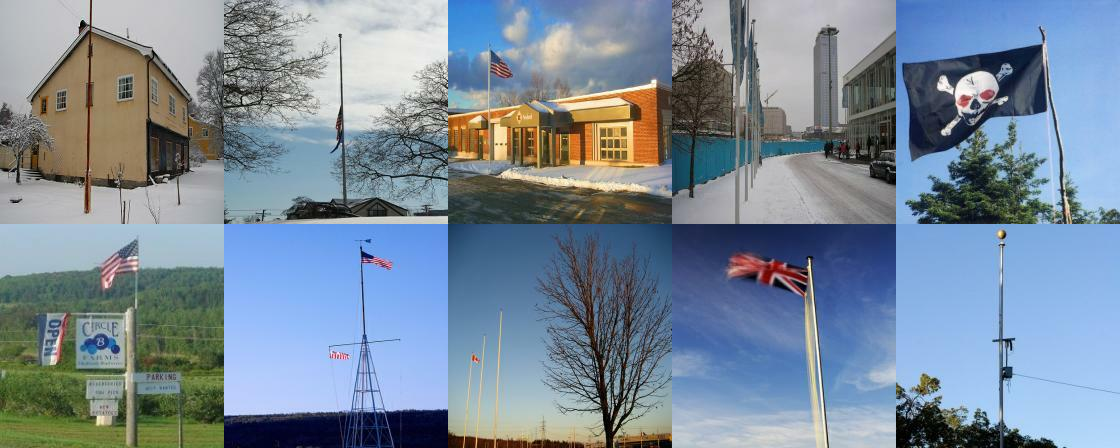}}
\caption{\textbf{Interpreting failure cases by inspecting their nearest neighbors in the train set of \imagenet.} We analyze the failure case in \ingvit shown in panel~\subref{fig:in_g_flag_ski}. The example is labeled as \texttt{flagpole} and classified as \texttt{ski} by a \vit trained on \imagenet. In panel~\subref{fig:nn_flag_ski_all} we show the 10 nearest neighbors of \subref{fig:in_g_flag_ski} in the train set of \imagenet. All 10 neighbors have the label \texttt{ski} and have similarities to the failure case, such as simultaneously having snow on the ground and blue sky in the background, while the 10 nearest neighbors with the label \texttt{flagpole} shown in \subref{fig:nn_flag_ski_same} do not present those common features. This suggests that the \vit correlates such features with instances from the label \texttt{ski}, and exploiting such spurious correlations likely induced it to misclassify the image in \subref{fig:in_g_flag_ski}.
}
\label{fig:nn_flag_as_ski}
\end{figure}

\autoref{fig:nn_mush_as_snail}-\ref{fig:nn_flag_as_ski} show \ingvit failure cases, along with their respective ten nearest neighbors within the \textit{full} \imagenet training set and the ten nearest neighbors with the same label.
Results suggest that failure cases found by our approach induce errors by generating images that have elements in their background which are more often found in other classes within \imagenet.
We further observe that all failure cases are closer to examples containing objects semantically related with cues present in images that are not commonly found in the training set of \imagenet for these classes.
In \autoref{fig:in_g_mush_snail}, for example, we show a failure case labeled as \texttt{mushroom} for which the \vit predicts the label \texttt{snail}.
All nearest neighbors shown in \autoref{fig:nn_mush_snail_all} are labeled as \texttt{snail} and contain elements such as human skin and grass in the background, which do not appear in the nearest neighbors from the label \texttt{mushroom}, as shown in \autoref{fig:nn_mush_snail_same}.
The \vit appears to be capturing spurious features in its representations (presence of human skin and grassy background) and relying on them to make predictions, which lead it to yield the wrong label for the \ingvit instance presented in \autoref{fig:in_g_mush_snail}.
In \autoref{fig:nn_butterfly_as_spider} and \ref{fig:nn_flag_as_ski}, we observe a similar pattern, where the \vit focuses on spurious cues such as the presence of a net in the background in \autoref{fig:in_g_butterfly_spider} and snow in \autoref{fig:in_g_flag_ski}.
Exploiting such correlations made the model mistake the particular instances of \texttt{cabbage butterfly} and \texttt{flagpole} as \texttt{barn spider} and \texttt{ski}, respectively.

\clearpage

\section{Malicious usage and mitigation strategies}

This work demonstrates how to find failure cases in vision classifiers with the help of large-scale generative models.
Much like adversarial examples~\citep{biggio_evasion_2013,szegedy_intriguing_2013}, malicious actors could leverage the proposed approach to build adversarial images that bypass automated online filtering mechanism.
In this section, we discuss how to make classifiers robust to these failure cases.

First, classifiers can be trained with discovered failure cases to make them more robust to generated inputs.
As a demonstration, we split the \ingvit dataset into a train and test set (80\% train, 20\% test).
We train the original \vit model in the exact same manner as before, except that batches are now made of 95\% \imagenet data and 5\% \ingvit data.
We report results with and without additional synthetic data in \autoref{tab:finetuneing}.
Training with additional generated data leads to a minimal loss of performance on \imagenet while achieving nearly 90\% top-1 accuracy on the \ingvit test set.
This demonstrates that adding the generated failure cases into the training set is an effective mitigation strategy.

Second, we note that our approach is computationally expensive.
It requires hundreds to thousands of calls to the generative model and vision classifier to find a {\em single} failure case.
Hiding the underlying classifier behind a rate-limited API can act as a first line of defense.

\begin{table}[h]
    \centering
    \small
    \begin{tabular}{ccc}
        Training Set & top-1 on \imagenet $\uparrow$ & top-1 on \ingvit $\uparrow$ \\ \toprule
        \imagenet (train) & \bf{82.57 $\pm$ 0.09} & 5.60 $\pm$ 2.80 \\
        \imagenet (train) + \ingvit (train) & 82.11 $\pm$ 0.05 & \bf{88.11 $\pm$ 0.44} \\ \bottomrule
    \end{tabular}
    \caption{{\bf top-1 accuracy on \imagenet and \ingvit}. We train a \vit model on either just \imagenet or \imagenet and \ingvit. By training on \ingvit, we achieve nearly 88\% top-1 accuracy on \ingvit (test) while minimally hurting performance on \imagenet. To obtain standard deviations, we run the experiment with $5$ random seeds.}
    \label{tab:finetuneing}
\end{table}

\clearpage

\section{Additional visualizations}

\begin{figure}[h]
\centering
\includegraphics[width=0.2\textwidth]{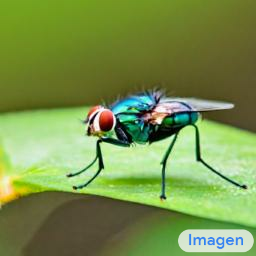}\hspace{.5cm}
\includegraphics[width=0.2\textwidth]{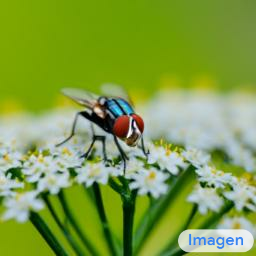}\hspace{.5cm}
\includegraphics[width=0.2\textwidth]{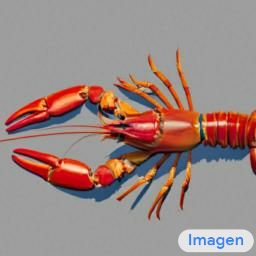}\hspace{.5cm}
\includegraphics[width=0.2\textwidth]{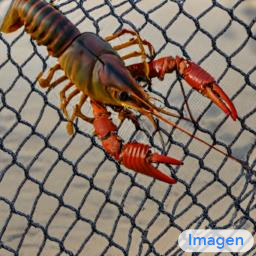}
\caption{\textbf{Images from the text-to-image model used in this manuscript.} Images are generated with captions identical to those used in \autoref{fig:openended_failures_fly} and \autoref{fig:openended_failures_crayfish}. A comparison with \dalle is shown in \autoref{fig:dalle}.}
\label{fig:imagen}
\end{figure}

\begin{figure}[h]
\centering
\includegraphics[width=0.2\textwidth]{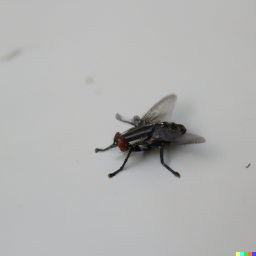}\hspace{.5cm}
\includegraphics[width=0.2\textwidth]{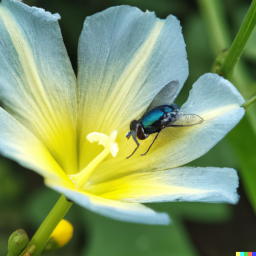}\hspace{.5cm}
\includegraphics[width=0.2\textwidth]{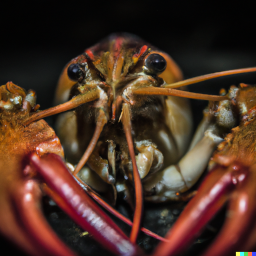}\hspace{.5cm}
\includegraphics[width=0.2\textwidth]{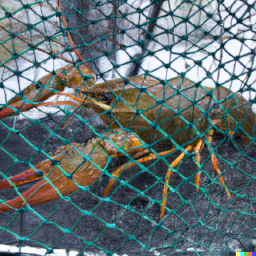}
\caption{\textbf{\dalle images.} Images are generated with captions identical to those used in \autoref{fig:openended_failures_fly} and \autoref{fig:openended_failures_crayfish}. A comparison with the text-to-image model used in the paper is shown in \autoref{fig:imagen}.}
\label{fig:dalle}
\end{figure}

\begin{figure}[h]
\centering
\includegraphics[width=0.2\textwidth]{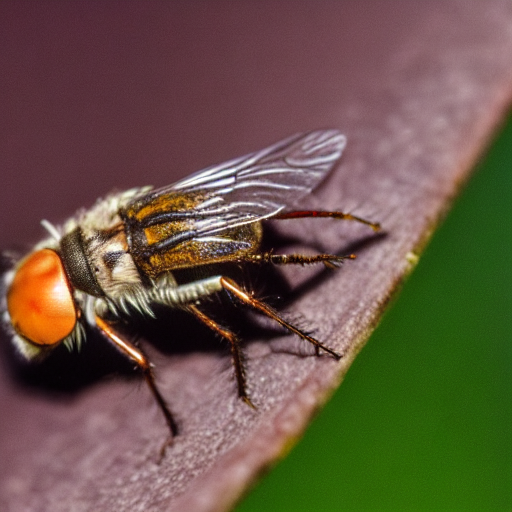}\hspace{.5cm}
\includegraphics[width=0.2\textwidth]{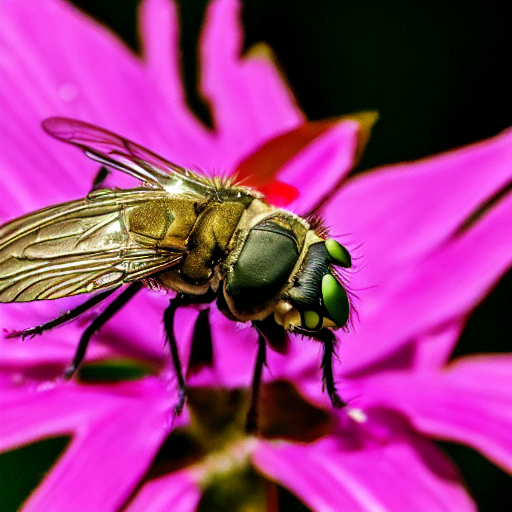}\hspace{.5cm}
\includegraphics[width=0.2\textwidth]{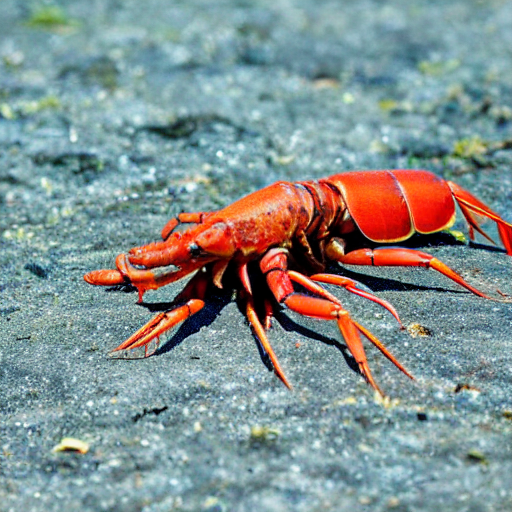}\hspace{.5cm}
\includegraphics[width=0.2\textwidth]{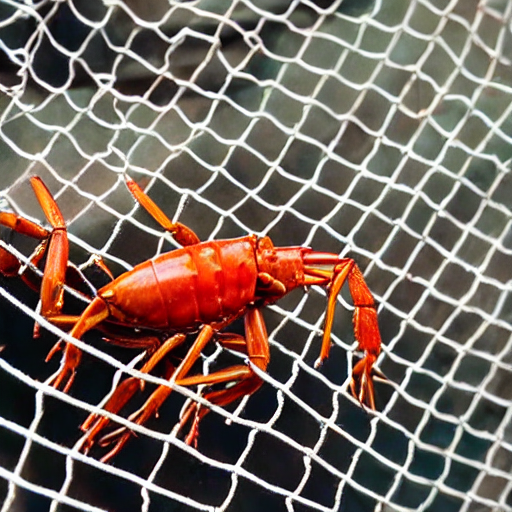}
\caption{\textbf{\stablediffusion images.} Images are generated with captions identical to those used in \autoref{fig:openended_failures_fly} and \autoref{fig:openended_failures_crayfish}. A comparison with the text-to-image model used in the paper is shown in \autoref{fig:imagen}.}
\label{fig:stablediffusion}
\end{figure}

\begin{figure}[b]
    \centering
    \begin{overpic}[width=\linewidth]{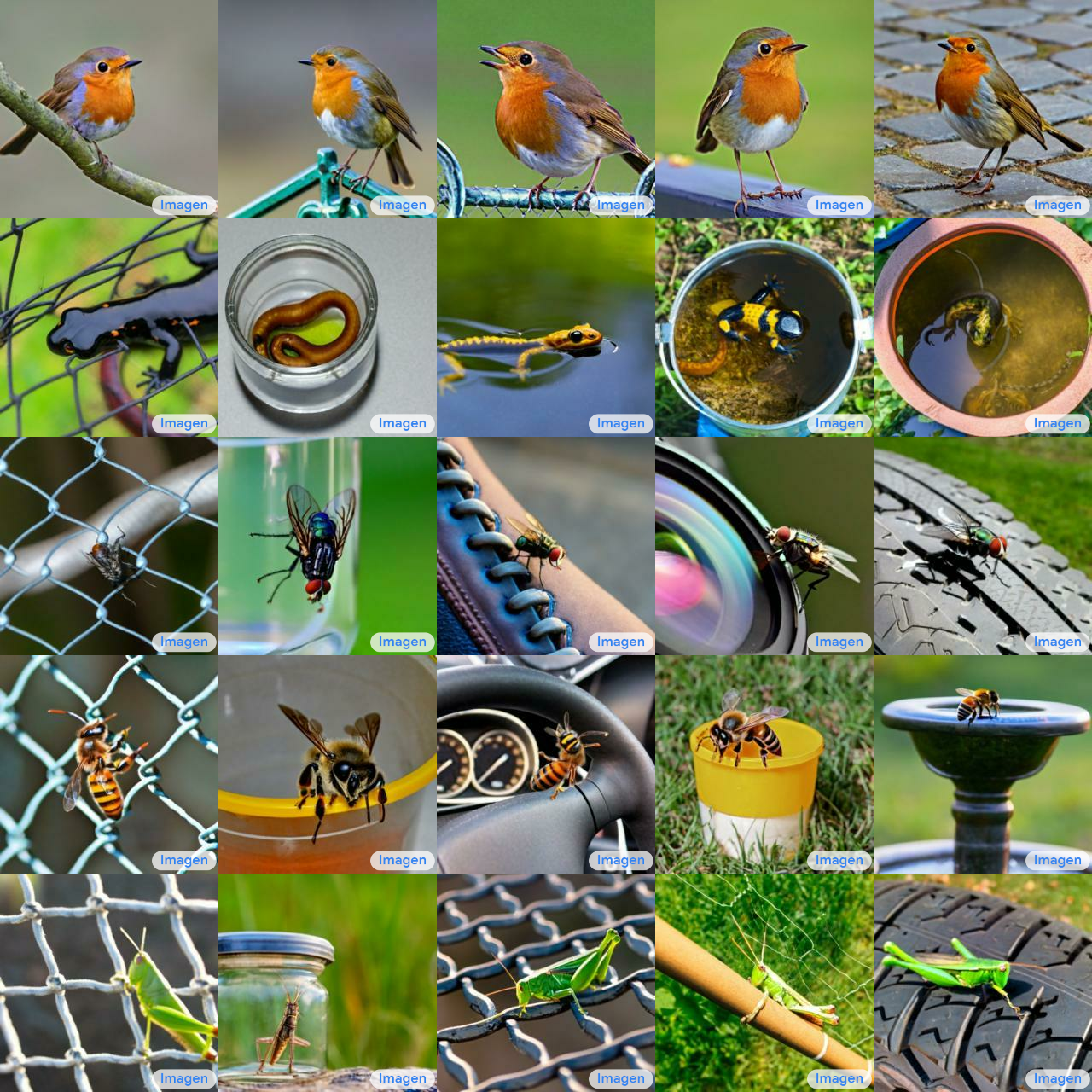}
    \put(10,101){\makebox[0pt]{\bf \color{red}{toucan}}}
    \put(30,101){\makebox[0pt]{\bf \color{red}{beaker}}}
    \put(50,101){\makebox[0pt]{\bf \color{red}{chain}}}
    \put(70,101){\makebox[0pt]{\bf \color{red}{reel}}}
    \put(90,101){\makebox[0pt]{\bf \color{red}{sundial}}}
    \put(-3,90){\rotatebox{90}{\makebox[0pt]{\bf \color{OliveGreen}{robin}}}}
    \put(-3,70){\rotatebox{90}{\makebox[0pt]{\bf \color{OliveGreen}{eft}}}}
    \put(-3,50){\rotatebox{90}{\makebox[0pt]{\bf \color{OliveGreen}{fly}}}}
    \put(-3,30){\rotatebox{90}{\makebox[0pt]{\bf \color{OliveGreen}{bee}}}}
    \put(-3,10){\rotatebox{90}{\makebox[0pt]{\bf \color{OliveGreen}{grasshopper}}}}
    \end{overpic}
    \caption{\textbf{Further examples from \ingresnet.} The label at the top of the column is one of the incorrectly predicted top-3 labels and the label on the left is the true label. }
    \label{fig:ingresnet}
\end{figure}

\begin{figure}[h]
    \centering
    \begin{overpic}[width=\linewidth]{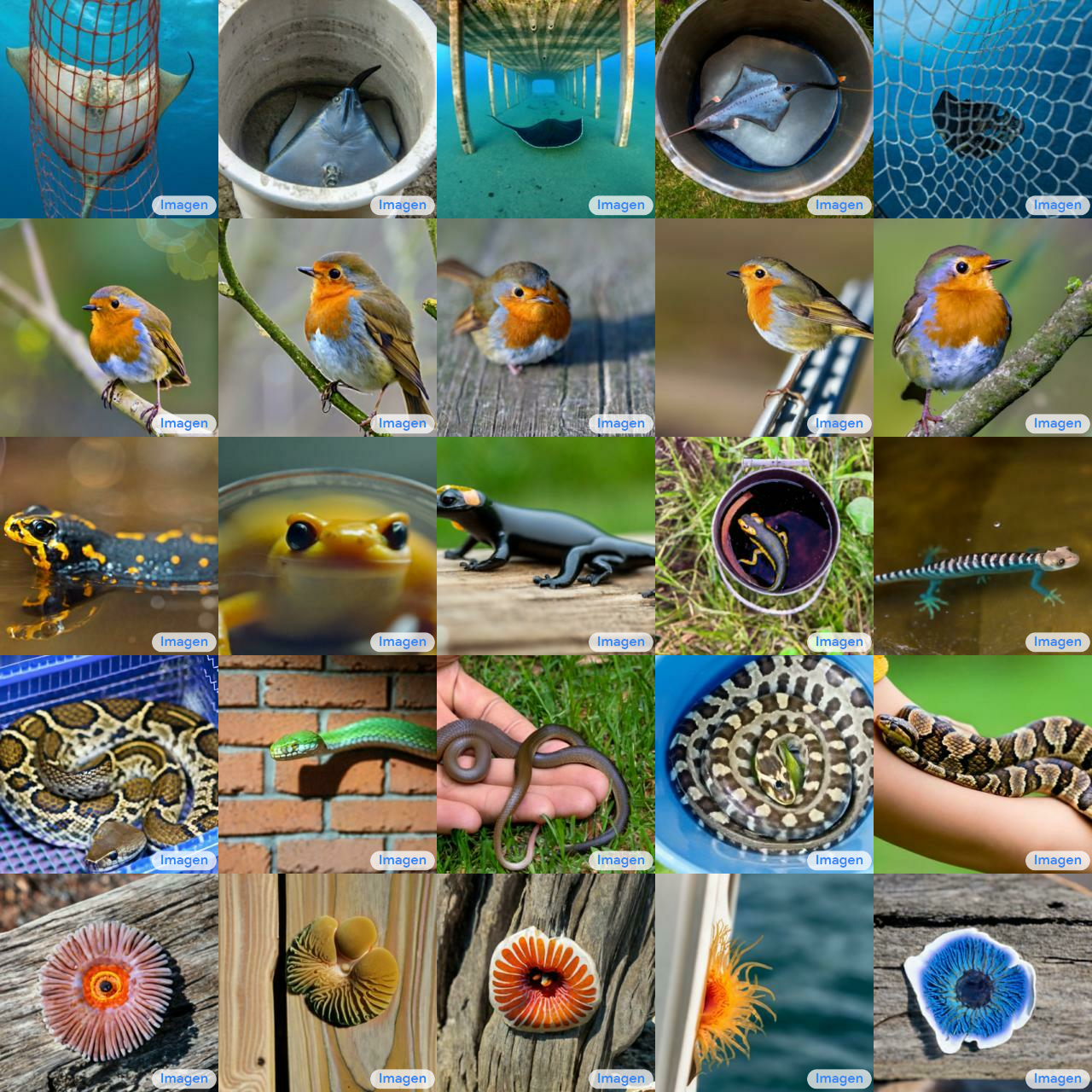}
    \put(10,101){\makebox[0pt]{\bf \color{red}{jellyfish}}}
    \put(30,101){\makebox[0pt]{\bf \color{red}{candle}}}
    \put(50,101){\makebox[0pt]{\bf \color{red}{nail}}}
    \put(70,101){\makebox[0pt]{\bf \color{red}{reel}}}
    \put(90,101){\makebox[0pt]{\bf \color{red}{bubble}}}
    \put(-3,90){\rotatebox{90}{\makebox[0pt]{\bf \color{OliveGreen}{stingray}}}}
    \put(-3,70){\rotatebox{90}{\makebox[0pt]{\bf \color{OliveGreen}{robin}}}}
    \put(-3,50){\rotatebox{90}{\makebox[0pt]{\bf \color{OliveGreen}{eft}}}}
    \put(-3,30){\rotatebox{90}{\makebox[0pt]{\bf \color{OliveGreen}{garter snake}}}}
    \put(-3,10){\rotatebox{90}{\makebox[0pt]{\bf \color{OliveGreen}{sea anemone}}}}
    \end{overpic}
    \caption{\textbf{Examples from \ingvit.} The label at the top of the column is one of the incorrectly predicted top-3 labels and the label on the left is the true label.}
    \label{fig:ingvit}
\end{figure}

\end{document}